\pdfoutput=1
\documentclass[10pt,twocolumn,letterpaper]{article}

\usepackage{cvpr}              
\pdfoutput=1
\usepackage{colortbl}
\definecolor{cvprblue}{rgb}{0.21,0.49,0.74}
\usepackage[pagebackref,breaklinks,colorlinks,allcolors=cvprblue]{hyperref}

\definecolor{blueColor}{HTML}{0054D6}
\PassOptionsToPackage{table,xcdraw, dvipsnames}{xcolor}
\usepackage[table,xcdraw]{xcolor}

\usepackage{url}
\usepackage{graphicx}
\usepackage{multirow}
\usepackage{pifont} 
\usepackage{booktabs}
\usepackage{makecell}
\usepackage{amsmath}
\usepackage{caption}
\usepackage{pifont}
\usepackage{amsmath}
\usepackage{threeparttable}
\usepackage{tcolorbox}
\usepackage[section]{placeins}
\usepackage{hyperref}  
\usepackage{adjustbox} 

\def\paperID{4350} 
\def\confName{CVPR}
\def\confYear{2025}

\usepackage{xspace}
\newcommand{\ours}{\text{SlideChat}\xspace}
\newcommand{\dataset}{\text{SlideInstruction}\xspace}

\title{SlideChat: A Large Vision-Language Assistant for  Whole-Slide Pathology Image Understanding}

\author{
Ying Chen$^{1,2*}$ \quad 
Guoan Wang$^{1,3*}$ \quad 
Yuanfeng Ji$^{4*}$\footnotemark[2] \quad
Yanjun Li$^{1,3}$ \quad
Jin Ye$^{1,5}$ \quad
Tianbin Li$^{1}$ \\
Ming Hu$^{1,5}$ \quad
Rongshan Yu$^{2}$ \quad
Yu Qiao$^{1}$ \quad
Junjun He$^{1}\footnotemark[2]$\\ \\
$^{1}$Shanghai AI Laboratory \quad 
$^{2}$Xiamen University \quad 
$^{3}$East China Normal University \\
$^{4}$Stanford University \quad
$^{5}$Monash University
}

\begin{document}
\maketitle

\let\thefootnote\relax\footnotetext{*These authors contributed equally to this work.}
\let\thefootnote\relax\footnotetext{\dag Corresponding authors: yfj@stanford.edu, hejunjun@pjlab.org.cn.}

\begin{abstract}
Despite the progress made by multimodal large language models (MLLMs) in computational pathology, they remain limited by a predominant focus on patch-level analysis, missing essential contextual information at the whole-slide level.
The lack of large-scale instruction datasets and the gigapixel scale of whole slide images (WSIs) pose significant developmental challenges. 
In this paper, we present SlideChat, the first vision-language assistant capable of understanding gigapixel whole-slide images, exhibiting excellent multimodal conversational capability and response complex instruction across diverse pathology scenarios.
To support its development, we created SlideInstruction, the largest instruction-following dataset for WSIs consisting of 4.2K WSI captions and 176K VQA pairs with multiple categories.
Furthermore, we propose SlideBench, a multimodal benchmark that incorporates captioning and VQA tasks to assess SlideChat's capabilities in various settings such as microscopy, diagnosis and clinical.
Compared to both general and specialized MLLMs, SlideChat exhibits exceptional  capabilities, achieving state-of-the-art performance on 18 of 22 tasks. 
For example, it achieved an overall accuracy of 81.17\% on SlideBench-VQA (TCGA), and 54.15\% on SlideBench-VQA (BCNB).
%
Our code, data, and model is publicly accessible at \href{https://uni-medical.github.io/SlideChat.github.io}{https://uni-medical.github.io/SlideChat.github.io.}

\end{abstract}    
\section{Introduction}
\label{sec:intro}

Computational pathology aims to improve the analysis of digitized tissue samples~\citep{hosseini2024computational,song2023artificial}, such as whole slide images (WSIs), by applying artificial intelligence to aid in the diagnosis, identification, and understanding of disease.
Recently, the development of this field has gained rapid momentum, mainly driven by breakthroughs in the visual foundation model~\citep{chen2024towards, xu2024whole, vorontsov2024foundation}.
These models learn generalized representations by pre-training on large-scale data and perform well in various downstream tasks, such as cancer detection and biomarker prediction.
Building on this base, integration with the powerful Large Language Models (LLMs) further advances the development of the Multimodal Large Language Model (MLLMs)~\citep{lu2024multimodal}, which has made great strides in responding to more complex, open-ended visual queries, enabling it to serve as a versatile assistant at various stages of medical care, including clinical decision-making, education, and research.

\begin{figure}[t]
  \centering
  \includegraphics[width=\linewidth]{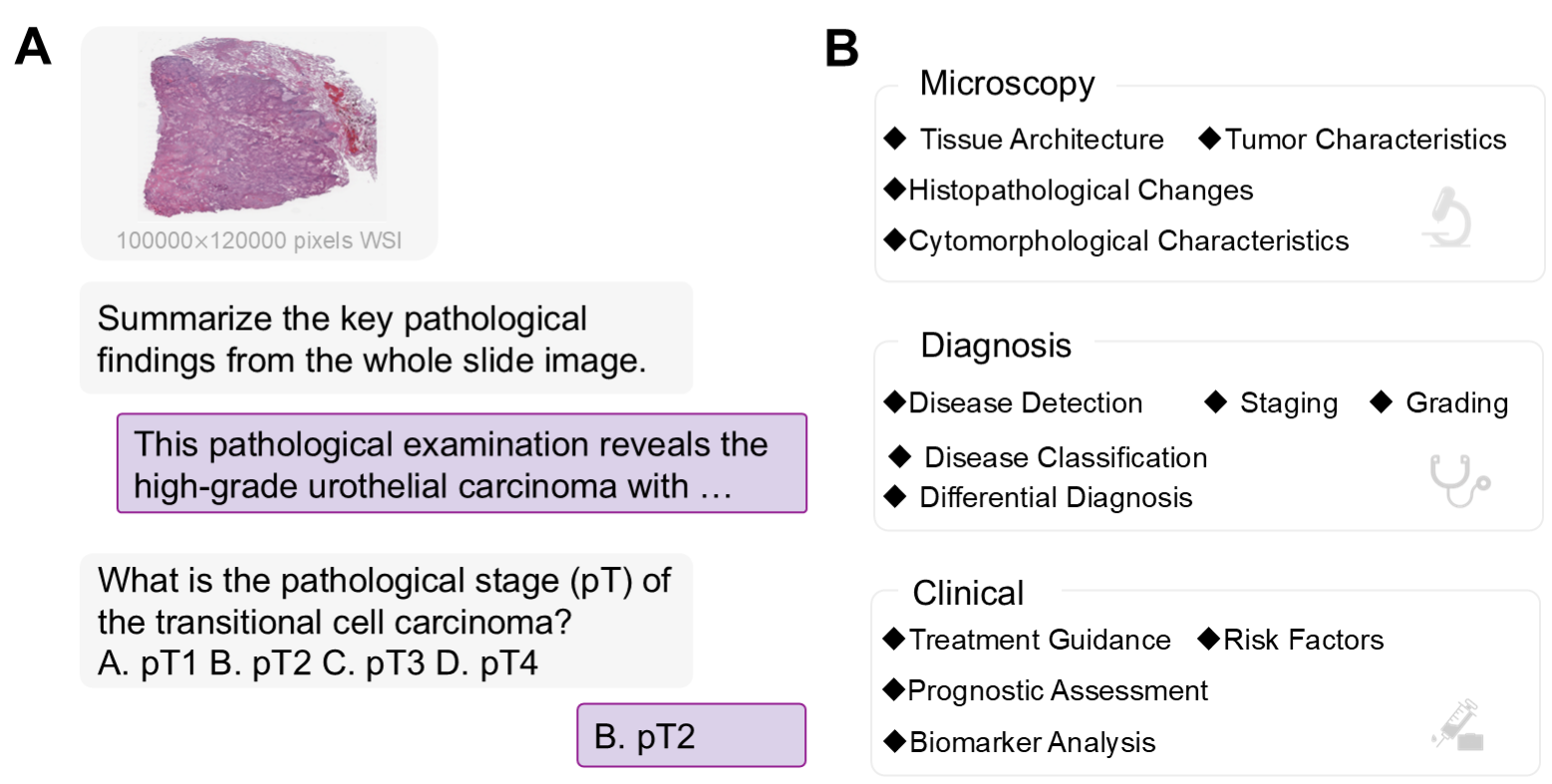}
  \caption{\ours is the first large vision-language assistant specifically designed for whole-slide pathology analysis. \ours can generate comprehensive descriptions of whole-slide images and provide contextually relevant responses across various applications.}
  \label{fig:SlideChat}
\end{figure}

Nevertheless, there are some major challenges that hinder the development and use of pathology MLLMs for real-world applications.
First, it is challenging to develop a MLLM architecture that can effectively be capable of gigapixel whole slides (\textit{e.g.}, 100,000 × 100,000 pixels). 
Existing models ~\citep{lu2024multimodal, sun2024pathasst, seyfioglu2024quilt} often process whole slides by extracting small patch/ROI-level data for subsequent analysis, resulting in a limited understanding of global slide context and suboptimal performance in some complex pathological analysis.
Second, publicly available multi-modal pathology slide datasets are relatively scarce and of varying quality~\citep{guo2024histgen, chen2023mi, chen2024wsi}, which limits the development of MLLMs trained on such data.

In this paper, we present \ours (see~\cref{fig:SlideChat}), the first open-source vision-language assistant capable of understanding gigapixel whole-slide images. 
First, \ours is trained on \dataset, a large-scale multi-modal instruction dataset encompassing data from The Cancer Genome Atlas (TCGA)~\citep{hutter2018cancer} via our specifically designed data processing pipeline (see \cref{fig:DataGen} (A)).
\dataset contains 4,181 WSI-caption pairs and 175,753 visual question-answer pairs from 3,294 patients, covering 10 cancer types.
The question-answer pairs include both open-ended and closed-ended questions, further divided into 13 subcategories, covering a diverse range of clinical tasks. 
\dataset is more than 20 times larger than previous public instruction datasets in the number of instructions.
Second, we propose \ours, a novel architecture in LLaVA style for capable multi-modal analysis of gigapixel whole slides.
A gigapixel whole slide is first divided into a series of patches, each of which is individually processed by a patch-level encoder to extract local features.
The resulting long sequence of patch tokens is then processed by a slide-level encoder employing sparse attention to enhance the slide-level features. 
Finally, enhanced patch tokens are fed into the LLM via a projector, which processes user queries and generates responses.

To systematically evaluate the performance of \ours in real-world scenarios, we establish a comprehensive digital pathology benchmark named SlideBench, encompassing more than 20 clinical tasks, using data from both TCGA and the in-the-wild Early Breast Cancer Core-Needle Biopsy (BCNB) dataset. 
This resulted in three test sets: SlideBench-Caption, comprising 734 WSI-caption pairs; SlideBench-VQA (TCGA), comprising 7,827 VQA pairs covering 13 different tasks; and SlideBench-VQA (BCNB), including a total of 7,274 VQA entries from 1,058 patients, covering seven different tasks. 
Additionally, we compare the performance of \ours on another externally proposed dataset, WSI-VQA~\citep{chen2024wsi}, to further validate its effectiveness.
We compare our model with the currently available state-of-the-art general and medical-specialized MLLMs including GPT-4o, LLava-Med~\citep{li2024llava}, MedDr~\citep{he2024meddr} and Quilt-LLaVA~\citep{seyfioglu2024quilt}. 
Benefiting from large-scale, high-quality training and effective local-global context modeling, \ours achieves state-of-the-art performance on 18 out of 22 tasks, with significant improvements over the second-best method 10\% on 9 tasks on four benchmarks.
Specifically, \ours achieves an average accuracy improvement of 13.47\% over the second-best model on SlideBench-VQA (TCGA), an average improvement of 12.59\% on SlideBench-VQA (BCNB), and an improvement of 5.82\% on WSI-VQA.
Finally, to accelerate research progress in digital pathology, we make \ours fully open-weight, including source code and model weights as well as instruction and benchmark datasets. 
The key contributions are summarized four-fold in the following:
\begin{itemize}[leftmargin=2em]
    \item We develop \ours, the first vision-language assistant capable of understanding gigapixel whole-slide images, achieving state-of-the-art performance on multiple benchmarks.
    \item We create \dataset, a largest comprehensive WSI instruction-following dataset containing 4.2K WSI-caption pairs and 176K VQA pairs. 
    \item We establish SlideBench, a WSI multi-modal benchmark comprising SlideBench-Caption, SlideBench-VQA (TCGA), and SlideBench-VQA (BCNB), covering 21 different clinical tasks.
    \item We release \ours, \dataset and SlideBench as open-source resources to facilitate research and development in computational pathology.
\end{itemize}
\section{Related Works}
\label{sec:Related Works}
\paragraph{Whole Slide Image Analysis}

Whole slide images are pivotal in modern pathology, enabling comprehensive analysis of tissue samples for tasks such as predicting patient prognosis, classifying cancer subtypes, and identifying biomarkers~\citep{song2023artificial, shao2023hvtsurv, li2024generalizable, spronck2023nnunet, lu2021ai, zhou2023cross}. Recent studies have leveraged pathology foundational models~\citep{wang2024pathology, ahmed2024pathalign, xu2024multimodal} to enhance WSIs analysis, either through fine-tuning for specific downstream tasks or by employing zero-shot prediction approaches in CLIP~\citep{radford2021learning} style. Although these models are effective in task-specific applications, their reliance on fine-tuning or limited zero-shot capabilities restricts their generalizability across diverse and complex user instructions.

\paragraph{MLLMs in Computational Pathology}
The paradigm of MLLMs enables them to effectively respond to more complex, open-ended visual queries while processing pathology images, thus providing significant value across various medical stages. 
PathChat~\citep{lu2024multimodal} is a vision-language assistant designed for pathology, developed with 450K private instruction pairs to handle both visual and natural language queries. 
Quilt-Instruct~\citep{seyfioglu2024quilt} is a large-scale dataset comprising 107K question-answer pairs.
Building on Quilt-Instruct, Quilt-LLAVA~\citep{seyfioglu2024quilt} is a model designed for diagnostic reasoning across multiple image patches, leveraging its extensive question-answer pairs to accurately interpret complex H\&E images. 
PathAsst~\citep{sun2024pathasst} combines a pathology-specific CLIP model with Vicuna-13b~\citep{chiang2023vicuna} to create a multimodal generative foundational model tailored for pathology.
However, current MLLMs primarily only focus on patch or ROI data, limiting their utility for slide-level clinical applications where broader contextual understanding is crucial.

\begin{figure*}[t]
  \centering
  \includegraphics[width=\linewidth]{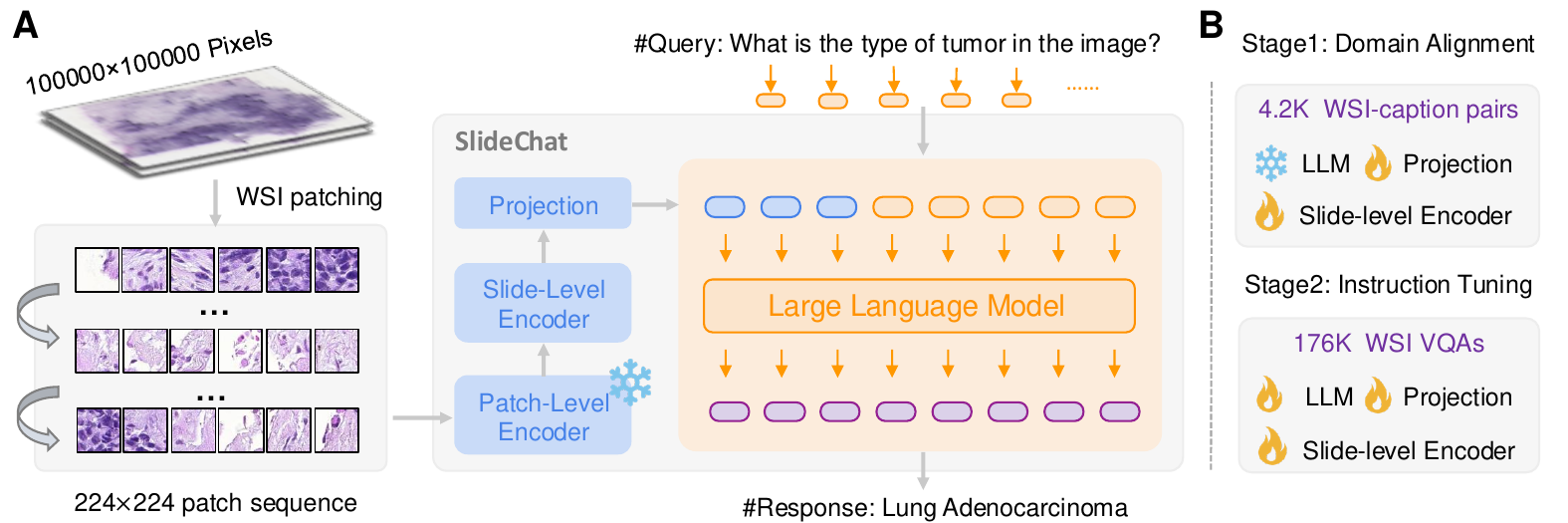}
  \caption{Overview of our SlideChat. (A) SlideChat serializes each input WSI into a sequence of 224×224 patches, converting each into visual embeddings with a patch-level encoder. A slide-level encoder then interacts with these features to generate contextual embeddings. Then, a multimodal projector maps the visual features from the slide-level encoder into a unified space, aligned seamlessly with the LLM. (B) SlideChat was trained for two stages: Cross-Domain Alignment and Visual Instruction Learning. }
  \label{fig:model}
\end{figure*}
\section{SlideChat}

\subsection{Architecture}
To achieve the goal of analyzing gigapixel whole-slide images in a multimodal setting, as shown in \cref{fig:model}, \ours consists of four key designs: the patch-level encoder, the slide-level encoder, the multimodal projector module, and the large language model. 
Our method starts by partitioning the WSI into smaller 224 × 224 pixel patches, making it computationally feasible to process such large images. 
These patches are then passed through a well-trained, frozen patch-level encoder~\citep{lu2024visual}, which extracts localized features from each individual patch, capturing fine-grained details such as cellular structures. 
Building on this, we employ LongNet~\citep{ding2023longnet,xu2024whole} as slide-level encoder to enhance the patch-level embeddings and capture global patterns across the entire slide. 
%
%
This encoder uses sparse attention mechanisms to learn both local and global contextual information, enabling the model to perceive intricate local features while capturing the broader context, which is critical for comprehensive pathological assessments.
Following the slide-level encoding, \ours incorporates a multimodal projector that maps these patch tokens into a unified space aligned with the LLM.
This ensures that the visual features extracted from the WSIs are effectively transformed into representations compatible with the language model, facilitating seamless integration and interaction between visual and textual data. 
Concurrently, the model accepts natural language instructions from users, such as “What is the type of tumor in the image?”.
These textual queries are processed by the LLM, which comprehends the textual input and integrates it with the visual features extracted from the WSIs, enabling accurate and contextually relevant diagnostic responses.
This multimodal reasoning capability allows SlideChat to provide accurate and contextually relevant answers to complex pathology-related questions, thereby supporting clinical decision-making, education, and research across various medical stages.

\begin{figure*}[t]
  \centering
  \includegraphics[width=\linewidth]{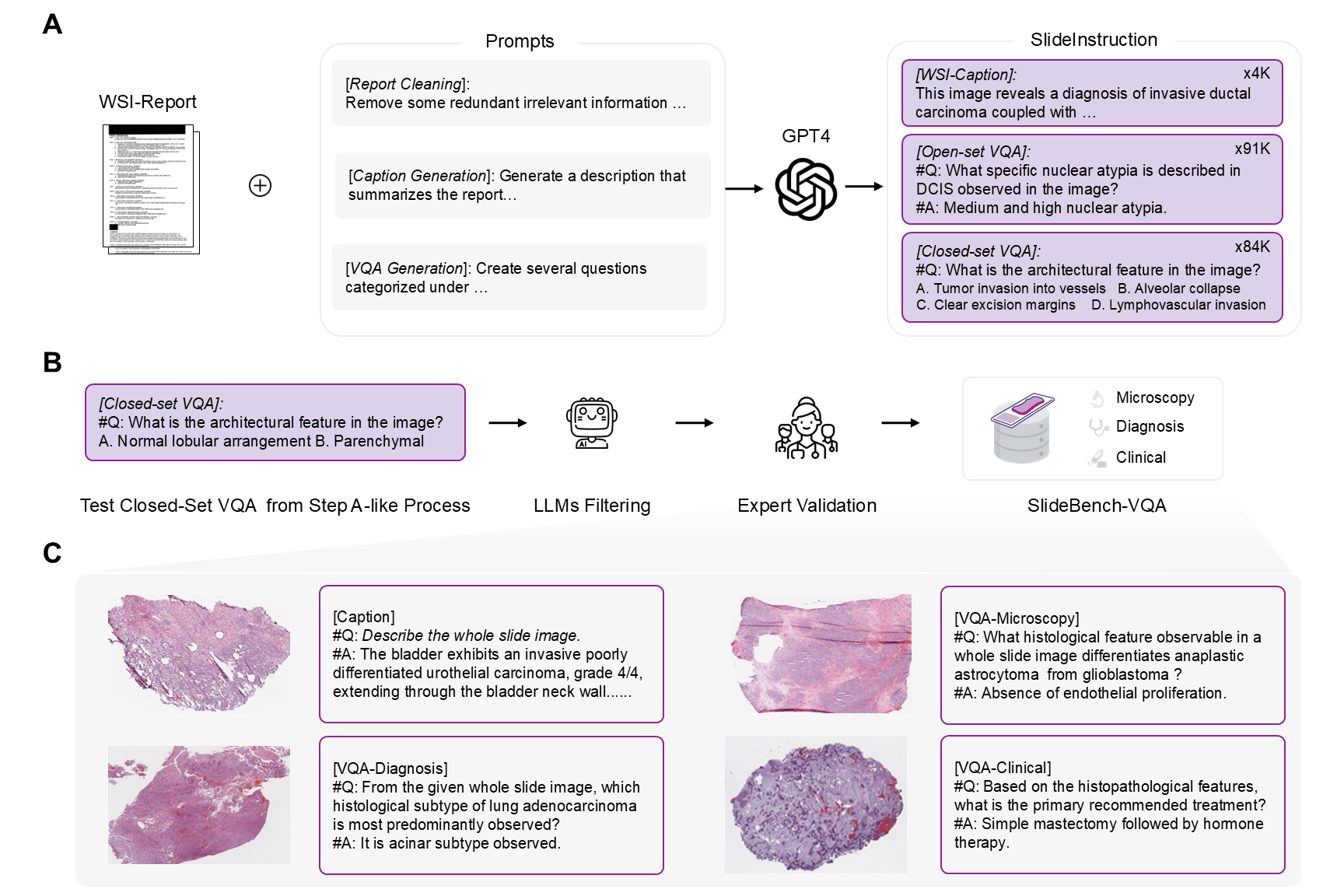}
  \caption{(A) Overview of the SlideInstruction generation pipeline. We prompt GPT-4 to extract the WSI-Caption, Open-set VQA and Closed-set VQA from reports. (B) For the generated Closed-set VQA, we employ LLMs to filter low-quality QA pairs and involve pathologists for validation, resulting in the creation of SlideBench-VQA. (C) Examples of WSI caption and instruction-following scenarios in microscopy, diagnostics, and clinical applications. For additional examples, please refer to Supplementary Material.}
  \label{fig:DataGen}
\end{figure*}

\subsection{Data}

\paragraph{SlideInstruction}

There is a notable lack of large-scale multimodal pathology datasets supporting the training of vision-language assistants for whole-slide image understanding. 
To support the training of \ours, we develop \dataset, a comprehensive instruction dataset, sourced from the TCGA database, comprising 4,915 WSI-report pairs from 4,028 patients.
\cref{fig:DataGen} illustrates our entire data curation pipeine. 
We initially prompt GPT-4 to refine the pathology reports, and clean up the noise in the report including \textit{unrelated symbols, technical details of pathology department procedures, specimen handling and processing information, redundant administrative or legal statements, and some repeated information}.
For the refined pathology reports, we further employ GPT-4 to generate high-quality multimodal data, comprising two main components: (1) \textit{WSI-Caption Data}: We craft concise, clinically relevant summaries for each whole slide image by prompting the language model to extract key pathological findings. 
These summaries were structured into coherent paragraphs that highlighted crucial clinical details such as diagnostic results, tumor characteristics, margin status, and lymph node involvement, ensuring the caption dataset is both focused and informative. 
(2) \textit{WSI Instruction-Following Data}: To enhance the model’s ability to follow instructions and improve its comprehension of pathology images, we leveraged GPT-4 to generate tailored question-and-answer pairs for each WSI report. 
Drawing inspiration from PathChat~\citep{lu2024multimodal}, we structure these questions into 3 broad categories—microscopy, diagnosis, and clinical considerations—which represent key stages in the pathology workflow, and 13 narrow categories focusing on specific aspects within each stage (\cref{fig:SlideChat} B). 
%
%
To construct a comprehensive instructional dataset, we began by generating two open-ended and two closed-ended QA pairs for each report within each specific category, taking advantage of the rich content found in the majority of TCGA reports. Then, some QA pairs were excluded, such as those lacking corresponding category information.
Regarding the train/test split, it is worth noting that the WSI-report datasets from TCGA include 2 types: (a) one report linked to multiple WSIs, and (b) one report linked to a single WSI.
For type (a), where specific diagnostic details may not align perfectly with each WSI, we include all WSIs in the training set to introduce some ``noisy data'', which can enhance model robustness. 
For type (b), 80\% of WSIs are allocated to the training set and 20\% to the test set. 
Finally, there are 4,181 WSIs for training and 734 WSIs for testing.
Consequently, we construct a large-scale training set named SlideInstruction, comprising 4,181 WSI captions and 175,753 instruction-following VQA pairs across 13 narrow categories.

\paragraph{SlideBench}
To systematically evaluate the performance of SlideChat, we include the remaining 734 WSI captions from the test set, along with a substantial number of closed-set VQA pairs, to establish a comprehensive evaluation benchmark.
First, we construct a set named SlideBench-Caption based on the WSI-Caption data to evaluate the model's ability to generate accurate and coherent descriptions of whole slide images.
Secondly, we construct SlideBench-VQA (TCGA) based on closed-set VQA pairs along with test WSIs, aiming to evaluate various aspects of model performance. 
As shown in \cref{fig:DataGen} (B), to improve the quality of the testing benchmarks, we employ four advanced large language models, including GPT-4~\citep{achiam2023gpt}, InternLM2-Chat-7B~\citep{cai2024internlm2}, Qwen-7B-Chat~\citep{bai2023qwen}, and DeepSeek-7B-Chat, to filter closed-set VQAs by predicting answers based solely on the question text. 
Any questions for which at least three of these models provided correct answers are subsequently excluded. 
Following this automated filtering, five expert pathologists are invited to review and amend the remaining questions. 
The review process is guided by the following criteria: (1) Whether the correct answer necessitates image interpretation; (2) Whether the question and its corresponding answer are logically and coherently structured; and (3) Whether the question aligns appropriately with the designated broad and narrow categories. 
VQA pairs failing to meet these criteria are excluded by the pathologists. 
Consequently, the SlideBench-VQA (TCGA) comprises 7,827 VQAs across 13 categories, with some examples illustrated in \cref{fig:DataGen} C.
Additionally, we incorporate the in-the-wild Early Breast Cancer Core-Needle Biopsy (BCNB) WSI dataset~\citep{xu2021predicting}, which encompasses a diverse patient population and a variety of clinical task labels, to enhance the test set benchmark and more comprehensively assess the model's generalization capabilities. 
In detail, we convert the BCNB dataset into a VQA format by rephrasing the classification objectives into a specific template as questions, while transforming the original multi-class labels into selectable options, and integrating it into SlideBench as an external subset, named SlideBench-VQA (BCNB).
This dataset comprises 7,247 VQA pairs from 1,058 patients, specifically designed to evaluate SlideChat's zero-shot generalization capability across 7 distinct classification tasks. 


\subsection{Two-stage Training}

\paragraph{Stage 1: Cross-Domain Alignment.}
\ours adopts a two-stage training approach (see \cref{fig:model} B). 
In the first stage, the primary objective is to align the LLM's word embeddings with the visual features extracted from whole slide images. 
This alignment enables the LLM to interpret visual representations from the slide-level encoder, facilitating the effective utilization of the intricate features within the slides. 
During this stage, SlideChat is trained to generate descriptive captions using 4.2K WSI-caption pairs from SlideInstruction. 
Specifically, only the slide-level encoder and projection matrix are updated, while the patch-level encoder and LLM weights remain fixed.

\paragraph{Stage 2: Visual Instruction Learning.} 
In the second stage, we focus on visual question-answering tasks to train the model to accurately respond to domain-specific questions concerning whole slide images.
During this phase, the model develops the ability to handle a broad range of multimodal instructions, enabling it to generate answers by effectively integrating both visual and textual information. 
For example, the model must perform various pathology tasks, such as describing the extent of tumor invasion or assessing the degree of cellular differentiation.
To accomplish this, we utilize 176K WSI VQAs from \dataset in the second training stage, allowing the slide encoder, projection layer, and large language model components to be fully trainable to ensure comprehensive adaptability. 
This approach significantly enhances the model’s capability to handle diverse pathology-related tasks, thereby increasing its effectiveness in real-world clinical and research settings.
\section{Experiment}
\label{sec:Experiment}

We conducted the following experiments to evaluate three key aspects of SlideChat: 
(1) its whole slide image captioning capability, which assesses proficiency in generating descriptive captions that accurately summarize the critical pathological features of a WSI; 
(2) its visual question-answering (VQA) ability across various complex pathological scenarios and its generalizability in zero-shot settings; 
and (3) SlideChat's ability to process gigapixel WSIs, capturing both essential global context and intricate details, thereby enhancing its performance compared to patch-level MLLMs.
For WSI captioning baselines, we benchmark against MI-Gen~\citep{chen2023mi}, a state-of-the-art method specifically designed for this task. Given that existing MLLMs cannot handle the gigapixel scale of whole slide images, we establish baseline comparisons using two approaches:
(1) randomly selecting 30 patches from each WSI and inputting them into MLLMs (e.g., GPT-4o~\citep{achiam2023gpt}, LLaVA-Med~\citep{li2024llava}, MedDr~\citep{li2024llava}, Quilt-LLaVA~\citep{seyfioglu2024quilt}), followed by a majority voting scheme to generate slide-level predictions; 
and (2) directly inputting a WSI thumbnail, resized to 1024×1024 pixels, referred to as Slide (T), into the MLLMs. 
For VQA tasks, we further evaluate performance by comparing against random prediction baselines and text-only models, thereby assessing the incremental contribution of visual information.
Unless otherwise specified, \ours is configured with the patch-level encoder CONCH~\citep{lu2024visual}, the  slide-level encoder LongNet~\citep{ding2023longnet}, and utilizes the Qwen2.5-7B-Instruct~\citep{yang2024qwen2} as LLM. 

\begin{table}[h]
\caption{Captioning performance across different methods on SlideBench-Caption.}
\label{tb: caption-baseline}
\centering
\setlength\tabcolsep{2.5pt}
\resizebox{\linewidth}{!}{
\begin{tabular}{ccccccccc}
\toprule
Methods  & Input &  BLEU-1 & BLEU-2 & BLEU-3 & BLEU-4 & Rouge-L & \makecell[c]{GPT\\score} \\ \hline
GPT-4o & Patch & 0.16	&	0.03	&	0.01	&	0.01	&	0.13	&	1.54 \\
GPT-4o & Slide (T) & 0.10	&	0.03	&	0.01	&	0.01	&	0.11	&	1.03 \\
Quilt-LLaVA & Patch & 0.30 &	0.19 &	0.11 &	0.05 &	0.18 &	3.87 \\
Quilt-LLaVA & Slide (T) & 0.23 &	0.09 &	0.04 &	0.01 &	0.16 &	1.89\\
MI-Gen & Slide & 0.37	&	0.24	&	0.15	&	0.10	&	0.25	&	4.14 \\ \hline
\rowcolor[HTML]{EFEFEF} 
SlideChat & Slide & 0.37	&	0.21	&	0.12	&	0.08	&	0.24	&	4.11 \\
\bottomrule
\end{tabular}
}
\end{table}

 \begin{table*}[t]
\caption{VQA performance with different methods. }
\label{tb: vqa-baseline}
\centering
\begin{tabular}{lcccccccc}
\toprule
\multirow{2}{*}{Methods}  & \multirow{2}{*}{Input} &  \multicolumn{4}{c}{SlideBench-VQA (TCGA)}  & \multirow{2}{*}{\makecell[c]{SlideBench-VQA\\(BCNB)}} & \multirow{2}{*}{\makecell[c]{WSI-VQA\textsuperscript{*}}}\\
\cmidrule(r){3-6} 
&  & Microscopy & Diagnosis & Clinical & Overall & ~ & ~\\ \hline
Random & \multirow{2}{*}{Text} &  24.44	&	24.91	&	26.44	&	25.02	&	24.40	&	24.14 \\
GPT-4 & ~ &  38.28	&	29.09	&	45.00	&	37.25	&	0	&	18.60 \\ 
\hline
GPT-4o & \multirow{4}{*}{Patch} &  62.89	&	46.69	&	66.77	&	57.91	&	41.43	&	30.41 \\
MedDr & ~  &    73.30	&	57.78	&	74.25	&	67.70	&	33.67	&	54.36 \\
LLaVA-Med & ~  &  47.34 &	32.78 &	47.96 &	42.00	& 30.1 &	26.31 \\
Quilt-LLaVA & ~  & 57.76 &	35.96 &	53.07 &	48.07 &	32.19 &	44.43\\
\hline
GPT-4o & \multirow{4}{*}{Slide (T)}  &  38.28	&	23.10	&	43.42	&	34.07	&	0	&	14.03\\
MedDr & ~  &  70.48	&	52.47	&	72.80	&	64.25	&	35.48	&	50.95 \\ 
LLaVA-Med & ~  &  45.82 &	27.58 &	40.84 &	37.39 &	0 &	18.79 \\
Quilt-LLaVA & ~  & 49.12 &	26.97 &	44.75 &	39.39 &	41.55 &	35.40\\
\hline
\rowcolor[HTML]{EFEFEF} 
SlideChat & Slide & \begin{tabular}[c]{@{}c@{}} 87.64 \\ \textcolor{blue}{(+14.34)} \end{tabular}	&	\begin{tabular}[c]{@{}c@{}} 73.27 \\ \textcolor{blue}{(+15.49)} \end{tabular}	&	\begin{tabular}[c]{@{}c@{}} 84.26 \\ \textcolor{blue}{(+10.01)} \end{tabular}	&	\begin{tabular}[c]{@{}c@{}} 81.17 \\ \textcolor{blue}{(+13.47)} \end{tabular}	&	\begin{tabular}[c]{@{}c@{}} 54.14	\\ \textcolor{blue}{(+12.59)} \end{tabular} &	 \begin{tabular}[c]{@{}c@{}}  60.18  \\ \textcolor{blue}{(+5.82)} \end{tabular} \\
\bottomrule
\end{tabular}
\end{table*}

\paragraph{SlideBench-Caption}

We report BLEU, ROUGE, and GPT scores to evaluate caption generation performance in ~\cref{tb: caption-baseline}. 
For the GPT score, we use GPT-4 to assess the similarity between the generated captions and the ground truth, providing an overall score on a scale of 1 to 10, with higher scores indicating better performance. 
When utilizing patch-level inputs, GPT-4o generates individual descriptions for each patch, which are subsequently integrated to create the final slide-level caption. 
However, this approach yields poor performance, as evidenced by a BLEU-1 score of 0.16 and a GPT-score of 1.54. These results suggest that the patch-based method fails to adequately capture the broader context necessary for accurate WSI captioning. 
When the WSI thumbnail of size 1024×1024 pixels is used as input to GPT-4o, performance decreases further, with a BLEU-1 score of 0.10 and a GPT-score of 1.03. 
This suggests that while the thumbnail offers a global view of the slide, it may lack the resolution and detail necessary for generating precise and informative captions.
In contrast, MI-Gen, a model specifically designed for WSI captioning, demonstrates significantly superior performance across all metrics, achieving a BLEU-1 score of 0.37 and a GPT score of 4.14.
Similarly, \ours, shows competitive results with a BLEU-1 score of 0.37 and a GPT score of 4.11. 
These outcomes highlight \ours's ability to effectively integrate both local and global information from the slides and confirm its efficiency in describing whole-slide images.

\begin{figure*}[ht]
  \centering
  \includegraphics[width=0.85\linewidth]{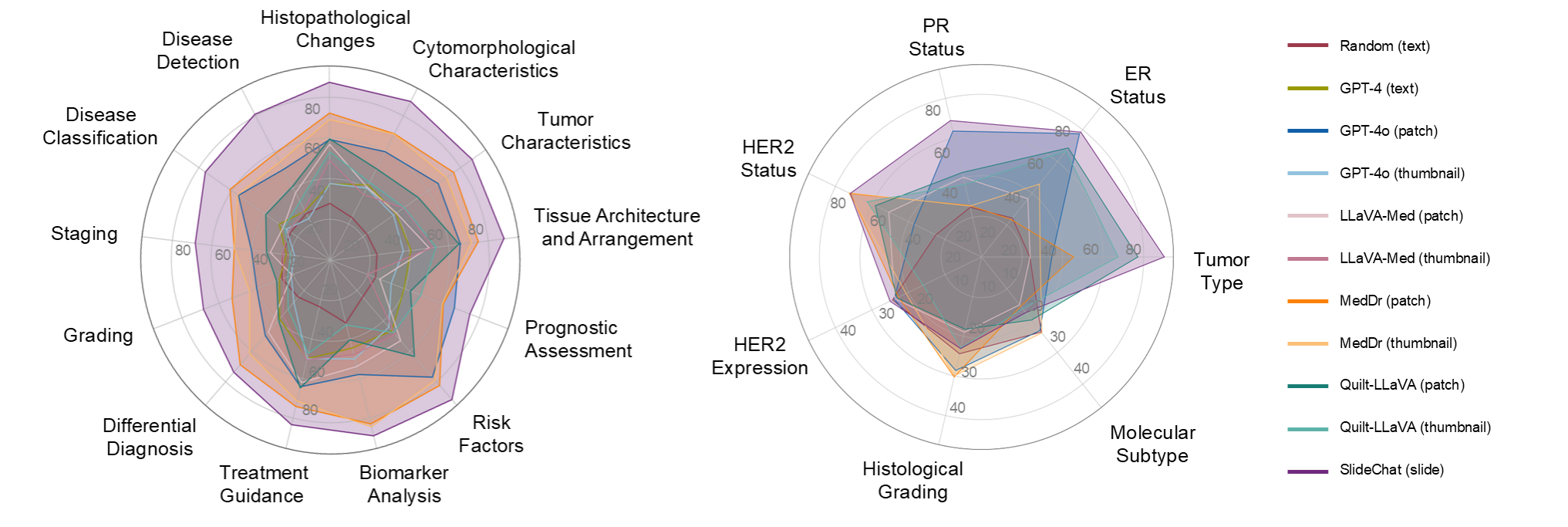}
  \caption{Accuracy on different tasks in SlideBench-VQA (TCGA) (left) and SlideBench-VQA (BCNB) (right).}
  \label{fig:radar}
\end{figure*}

\paragraph{SlideBench-VQA (TCGA)}
We further evaluate \ours's overall performance on the multiple-choice VQA benchmark. 
The results on SlideBench-VQA (TCGA) in~\cref{tb: vqa-baseline}, compare different methods across three categories: microscopy, diagnosis, and clinical, along with an overall performance score. 
Random selection achieves an overall score of 25.02\% accuracy, serving as a baseline for answer distribution but demonstrating poor performance. 
While GPT-4, relying solely on text input, outperforms random predictions, it continues to struggle with accurately answering questions.
When GPT-4o incorporates patch-level inputs, its performance improves markedly, reaching a score of 57.91\% and underscoring the crucial role of detailed visual data. 
However, using a WSI thumbnail results in a lower score of 34.07\%, as the reduced detail restricts its ability to deliver precise answers. 
MedDr performs well, achieving an overall score of 67.70\% with patch-level inputs, though this drops slightly to 64.25\% when using the slide thumbnail due to the loss of fine visual details.
SlideChat outperforms all other methods, attaining a leading overall accuracy of 81.17\%, excelling across all categories and significantly surpassing the competition.
Even in more fine-grained pathological scenarios, as depicted in the left portion of~\cref{fig:radar}, \ours remains the top-performing model across 13 tasks, particularly in areas such as cytomorphological characteristics, histopathological changes, disease detection, disease classification, and staging and grading, which require the identification of complex pathological visual features. 
Compared with baselines taking some patches or slide thumbnail as inputs, \ours has the capability to analyze a significantly greater number of pathological features with enhanced detail, effectively capturing both localized features and overarching global patterns, allowing \ours to provide more accurate and nuanced insights into pathological variations.
%
%

\paragraph{SlideBench-VQA (BCNB)} 

%
In the zero-shot VQA setting, SlideChat significantly outperforms all other models, achieving the highest overall score of 54.14\% on SlideBench-VQA (BCNB). 
It particularly excels in identifying tumor types, far surpassing other baselines in this task. This performance highlights SlideChat's generalization capability across a wide range of tasks.
When taking patches as inputs, GPT-4o outperforms both MedDr and LLaVA-Med, achieving a score of 41.43\%, though it still falls short of SlideChat by 12.59\%. Notably, GPT-4o and LLaVA-Med performed very poorly, achieving a score of zero across all tasks when evaluated using slide thumbnails from this testing set.
MedDr and Quilt-LLaVA both achieve moderate results, with MedDr scoring 33.67\% on patches and 35.48\% on thumbnails, while Quilt-LLaVA attains 32.19\% on patches and 41.55\% on thumbnails.
%
This outcome highlights that, for complex WSIs, relying solely on relatively sufficient visual features is inadequate for effectively supporting a range of tasks.
In the more fine-grained pathological tasks of the BCNB benchmark, \ours attains state-of-the-art performance on 5 out of 7 tasks (\cref{fig:radar}), further demonstrating the effectiveness of \ours.

\paragraph{WSI-VQA\textsuperscript{*}}
We also curated a subset of closed-set VQA pairs from the public WSI-VQA~\citep{chen2024wsi} dataset, referred to as WSI-VQA\textsuperscript{*}, which exclusively consists of samples that were not in the training set, to evaluate the model's performance.
SlideChat demonstrates the highest performance with a score of 60.18. Although MedDr performs well with both patch inputs (54.36\%) and slide thumbnail inputs (50.95\%), it still falls short compared to SlideChat. 
GPT-4o struggles significantly, especially with slide thumbnails, scoring only 14.03\%, which highlights the limitations of using lower-resolution inputs. 
SlideChat leverages both detailed local information and broader context to process WSIs, making it the most effective model for this benchmark.
This further emphasizes its superior capability in handling whole-slide data for VQA tasks.

\begin{table}[h]
\caption{Performance Comparison of LLMs and Slide Encoder on WSI Captioning and VQA Tasks.}
\label{tb: caption different LLMs}
\begin{center}
\setlength\tabcolsep{2.5pt}
\resizebox{\linewidth}{!}{
\begin{tabular}{lccccc}
\toprule
LLMs & \makecell[c]{Slide\\Encoder} &  Caption & \makecell[c]{VQA\\(TCGA)} & \makecell[c]{VQA\\(BCNB)} & WSI-VQA\textsuperscript{*}\\ \hline
Vicuna-7B-v1.5 & \checkmark & 3.28 &	41.43	&	41.43	&	31.98 \\
Phi-3-Mini-4k-Instruct & \checkmark & 2.66 &	79.93	&	43.92	&	60.18 \\
Qwen1.5-7B-Chat & \checkmark &	2.92 &	77.63	&	44.07	&	56.89 \\ 
Llama3-8B-Instruct & \checkmark &	3.30 &	78.78	&	42.82	&	55.25 \\
Internlm2-Chat-7B & \checkmark &	3.30  &	79.10	&	52.13	&	56.76 \\ \hline
Qwen2.5-3B-Instruct & \checkmark &	3.40 &	80.32 &	45.79 &	56.38 \\
Qwen2.5-14B-Instruct & \checkmark &	3.39 &	82.14 &	51.57 &	60.94  \\
\rowcolor[HTML]{EFEFEF} 
Qwen2.5-7B-Instruct & \checkmark &	4.11 &	81.17	&	54.14	&	60.18 \\
Qwen2.5-7B-Instruct & $\times$ & 3.95 &	81.21	&	45.49	&	59.67 \\
\bottomrule
\end{tabular}
}
\end{center}
\end{table}

\begin{figure*}[h]
  \centering
  \includegraphics[width=0.8\linewidth]{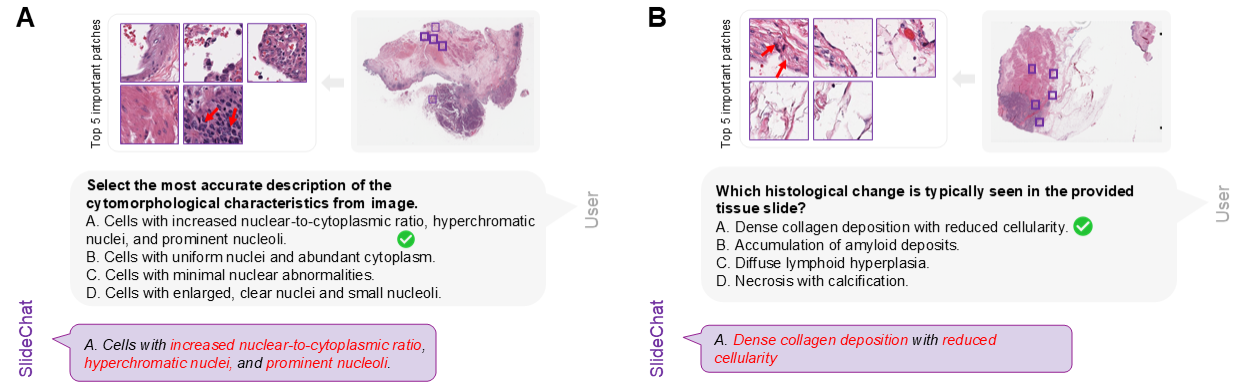}
  \caption{Interpretability and visualization. We identify the top five patch tokens with the highest attention scores associated with the output text responses.}
  \label{fig:Interpretability}
\end{figure*}

\paragraph{Ablation}
We performed ablation experiments from different perspectives as follows:
a) \textit{Large Language Model Comparison}: Firstly, we compare the performance of several large language models, each with a parameter scale of approximately 7 billion, including Vicuna-7B-v1.5~\citep{chiang2023vicuna}, Phi-3-Mini-4k-Instruct~\citep{abdin2024phi}, Qwen1.5-7B-Chat~\citep{bai2023qwen}, Llama3-8B-Instruct~\citep{llama3modelcard}, InternLM2-Chat-7B~\citep{cai2024internlm2}, and Qwen2.5-7B-Instruct~\citep{yang2024qwen2}. 
Specifically, we measured their performance on the SlideBench-Caption using the GPT-score and their accuracy on three VQA datasets. 
As shown in ~\cref{tb: caption different LLMs}, the results demonstrate that SlideChat, powered by the Qwen2.5-7B-Instruct model, achieved the highest performance across all tasks, particularly excelling in the captioning task.
These findings underscore the significant potential of developing SlideChat with the Qwen2.5-7B-Instruct model. 
Utilizing the Qwen2.5 model, we further evaluated models of varying scales and discovered that larger models generally exhibited superior performance, particularly the 7B and 14B parameter models. 
While these two models showed comparable performance across the three benchmarks, the 14B model surpassed the 7B model by 2.57\% on the SlideBench-VQA (TCGA). 
Given computational resource constraints, SlideChat uses the 7B model by default to achieve the best balance between performance and resource efficiency.
b) \textit{Slide-level encoder effectiveness}: We investigate the effectiveness of the slide-level encoder by initially removing it from SlideChat and employing a two-stage training approach. In the first stage, we exclusively trained the projection layers. However, this approach failed to reduce training loss or generate coherent text effectively, likely due to the difficulty of learning the complex visual features of WSIs without the slide-level encoder. When using a frozen LLM for complex text generation, a simple projection proved inadequate for effectively integrating and aligning visual and textual features.
Subsequently, we consider combining data from both stages and training SlideChat without the slide-level encoder by simultaneously updating both the projection layers and the LLM. Under this paradigm, performance on SlideBench-VQA (TCGA) and WSI-VQA\textsuperscript{*}, which share the same distribution as the training set, was comparable to the two-stage training configuration with the slide-level encoder. 
However, a significant decline is observed when evaluating SlideBench-VQA (BCNB), which originates from a different domain; overall performance dropped by over 10\% (\cref{tb: caption different LLMs}), indicating a substantial reduction in the model's generalization ability. 
Therefore, we recommend incorporating a slide-level encoder to capture the complex visual features of whole slide, as it is particularly effective for cross-domain alignment and enhances the model's generalization performance.

\paragraph{Interpretability}

Despite SlideChat demonstrating promising results, concerns remained regarding the model's perception of large pathological slides. 
To further assess the model's interpretability, we calculated the correlation between the text output and specific image patches, thereby obtaining patch-level attention scores.
By identifying the most significant patches, we gained insights into the precise areas the model focused on during response generation. 
Highlighting the most relevant regions of the tissue slides not only enhances transparency and bolsters the reliability of the model’s outputs but also assists pathologists by directing attention to critical areas requiring closer scrutiny. 
Ultimately, such interpretability is essential for fostering trust in AI-assisted diagnostics and enhancing the precision and efficiency of clinical evaluations.
As shown in ~\cref{fig:Interpretability}, we are pleased to observe that the top five important patches identified by the model closely corresponded with the features described in SlideChat's output. 
Our extraction method retrieves attention weights for patch tokens from each generated token, averaging them across layers and heads. We then identify the top five patch tokens with the highest attention weights for further analysis.
For example, in \cref{fig:Interpretability} (A), the highlighted patches clearly emphasized regions exhibiting an increased nuclear-to-cytoplasmic ratio, hyperchromatic nuclei, and prominent nucleoli. 
Similarly, in \cref{fig:Interpretability} (B), the selected patches demonstrated areas with dense collagen deposition and reduced cellularity, as detailed in the model’s response. 
This alignment between the highlighted image regions and the textual outputs significantly enhances the model's interpretability, providing increased confidence that it accurately captures and assesses relevant histopathological features. 
Such consistency deepens our understanding of the model's reasoning processes regarding pathological slides and underscores the potential for integrating these AI systems into clinical workflows with greater assurance.


\section{Conclusion}
\label{sec:Conclusion}

In this work, we present SlideChat, the first vision-language assistant capable of understanding gigapixel whole-slide images. Furthermore, we creat SlideInstruction, a largest comprehensive WSI instruction-following dataset to develop SlideChat, as well as SlideBench, a multi-modal benchmark designed to evaluate SlideChat across diverse scenarios. SlideChat demonstrates excellent chat abilities and achieves state-of-the-art performance on 18 tasks.
We bridge the gap between MLLMs and pathology images at the whole-slide level with SlideChat, and we believe this represents an advancement in the field of general medical artificial intelligence (GMAI).

\paragraph{Acknowledgement} The authors acknowledge the funding provided by the National Key Research and Development Program of China (No.2022ZD0160102). 
{
    \small
    \bibliographystyle{ieeenat_fullname}
    \bibliography{reference}

\begin{thebibliography}{35}
\providecommand{\natexlab}[1]{#1}
\providecommand{\url}[1]{\texttt{#1}}
\expandafter\ifx\csname urlstyle\endcsname\relax
  \providecommand{\doi}[1]{doi: #1}\else
  \providecommand{\doi}{doi: \begingroup \urlstyle{rm}\Url}\fi

\bibitem[Abdin et~al.(2024)Abdin, Jacobs, Awan, Aneja, Awadallah, Awadalla, Bach, Bahree, Bakhtiari, Behl, et~al.]{abdin2024phi}
Marah Abdin, Sam~Ade Jacobs, Ammar~Ahmad Awan, Jyoti Aneja, Ahmed Awadallah, Hany Awadalla, Nguyen Bach, Amit Bahree, Arash Bakhtiari, Harkirat Behl, et~al.
\newblock Phi-3 technical report: A highly capable language model locally on your phone.
\newblock \emph{arXiv preprint arXiv:2404.14219}, 2024.

\bibitem[Achiam et~al.(2023)Achiam, Adler, Agarwal, Ahmad, Akkaya, Aleman, Almeida, Altenschmidt, Altman, Anadkat, et~al.]{achiam2023gpt}
Josh Achiam, Steven Adler, Sandhini Agarwal, Lama Ahmad, Ilge Akkaya, Florencia~Leoni Aleman, Diogo Almeida, Janko Altenschmidt, Sam Altman, Shyamal Anadkat, et~al.
\newblock Gpt-4 technical report.
\newblock \emph{arXiv preprint arXiv:2303.08774}, 2023.

\bibitem[Ahmed et~al.(2024)Ahmed, Sellergren, Yang, Xu, Babenko, Ward, Olson, Mohtashamian, Matias, Corrado, et~al.]{ahmed2024pathalign}
Faruk Ahmed, Andrew Sellergren, Lin Yang, Shawn Xu, Boris Babenko, Abbi Ward, Niels Olson, Arash Mohtashamian, Yossi Matias, Greg~S Corrado, et~al.
\newblock Pathalign: A vision-language model for whole slide images in histopathology.
\newblock \emph{arXiv preprint arXiv:2406.19578}, 2024.

\bibitem[AI@Meta(2024)]{llama3modelcard}
AI@Meta.
\newblock Llama 3 model card.
\newblock 2024.

\bibitem[Bai et~al.(2023)Bai, Bai, Chu, Cui, Dang, Deng, Fan, Ge, Han, Huang, et~al.]{bai2023qwen}
Jinze Bai, Shuai Bai, Yunfei Chu, Zeyu Cui, Kai Dang, Xiaodong Deng, Yang Fan, Wenbin Ge, Yu Han, Fei Huang, et~al.
\newblock Qwen technical report.
\newblock \emph{arXiv preprint arXiv:2309.16609}, 2023.

\bibitem[Cai et~al.(2024)Cai, Cao, Chen, Chen, Chen, Chen, Chen, Chen, Chen, Chu, et~al.]{cai2024internlm2}
Zheng Cai, Maosong Cao, Haojiong Chen, Kai Chen, Keyu Chen, Xin Chen, Xun Chen, Zehui Chen, Zhi Chen, Pei Chu, et~al.
\newblock Internlm2 technical report.
\newblock \emph{arXiv preprint arXiv:2403.17297}, 2024.

\bibitem[Chen et~al.(2023)Chen, Li, Zhu, Zheng, and Yang]{chen2023mi}
Pingyi Chen, Honglin Li, Chenglu Zhu, Sunyi Zheng, and Lin Yang.
\newblock Mi-gen: Multiple instance generation of pathology reports for gigapixel whole-slide images.
\newblock \emph{arXiv preprint arXiv:2311.16480}, 2023.

\bibitem[Chen et~al.(2024{\natexlab{a}})Chen, Zhu, Zheng, Li, and Yang]{chen2024wsi}
Pingyi Chen, Chenglu Zhu, Sunyi Zheng, Honglin Li, and Lin Yang.
\newblock Wsi-vqa: Interpreting whole slide images by generative visual question answering.
\newblock \emph{arXiv preprint arXiv:2407.05603}, 2024{\natexlab{a}}.

\bibitem[Chen et~al.(2024{\natexlab{b}})Chen, Ding, Lu, Williamson, Jaume, Song, Chen, Zhang, Shao, Shaban, et~al.]{chen2024towards}
Richard~J Chen, Tong Ding, Ming~Y Lu, Drew~FK Williamson, Guillaume Jaume, Andrew~H Song, Bowen Chen, Andrew Zhang, Daniel Shao, Muhammad Shaban, et~al.
\newblock Towards a general-purpose foundation model for computational pathology.
\newblock \emph{Nature Medicine}, 30\penalty0 (3):\penalty0 850--862, 2024{\natexlab{b}}.

\bibitem[Chiang et~al.(2023)Chiang, Li, Lin, Sheng, Wu, Zhang, Zheng, Zhuang, Zhuang, Gonzalez, et~al.]{chiang2023vicuna}
Wei-Lin Chiang, Zhuohan Li, Zi Lin, Ying Sheng, Zhanghao Wu, Hao Zhang, Lianmin Zheng, Siyuan Zhuang, Yonghao Zhuang, Joseph~E Gonzalez, et~al.
\newblock Vicuna: An open-source chatbot impressing gpt-4 with 90\%* chatgpt quality.
\newblock \emph{See https://vicuna. lmsys. org (accessed 14 April 2023)}, 2\penalty0 (3):\penalty0 6, 2023.

\bibitem[Contributors(2023)]{contributors2023xtuner}
XTuner Contributors.
\newblock Xtuner: A toolkit for efficiently fine-tuning llm, 2023.

\bibitem[Ding et~al.(2023)Ding, Ma, Dong, Zhang, Huang, Wang, Zheng, and Wei]{ding2023longnet}
Jiayu Ding, Shuming Ma, Li Dong, Xingxing Zhang, Shaohan Huang, Wenhui Wang, Nanning Zheng, and Furu Wei.
\newblock Longnet: Scaling transformers to 1,000,000,000 tokens.
\newblock \emph{arXiv preprint arXiv:2307.02486}, 2023.

\bibitem[Guo et~al.(2024)Guo, Ma, Xu, Wang, Wang, and Chen]{guo2024histgen}
Zhengrui Guo, Jiabo Ma, Yingxue Xu, Yihui Wang, Liansheng Wang, and Hao Chen.
\newblock Histgen: Histopathology report generation via local-global feature encoding and cross-modal context interaction.
\newblock \emph{arXiv preprint arXiv:2403.05396}, 2024.

\bibitem[He et~al.(2024)He, Nie, Chen, Cai, Wang, Yang, and Chen]{he2024meddr}
Sunan He, Yuxiang Nie, Zhixuan Chen, Zhiyuan Cai, Hongmei Wang, Shu Yang, and Hao Chen.
\newblock Meddr: Diagnosis-guided bootstrapping for large-scale medical vision-language learning.
\newblock \emph{arXiv preprint arXiv:2404.15127}, 2024.

\bibitem[Hosseini et~al.(2024)Hosseini, Bejnordi, Trinh, Chan, Hasan, Li, Yang, Kim, Zhang, Wu, et~al.]{hosseini2024computational}
Mahdi~S Hosseini, Babak~Ehteshami Bejnordi, Vincent Quoc-Huy Trinh, Lyndon Chan, Danial Hasan, Xingwen Li, Stephen Yang, Taehyo Kim, Haochen Zhang, Theodore Wu, et~al.
\newblock Computational pathology: a survey review and the way forward.
\newblock \emph{Journal of Pathology Informatics}, page 100357, 2024.

\bibitem[Hutter and Zenklusen(2018)]{hutter2018cancer}
Carolyn Hutter and Jean~Claude Zenklusen.
\newblock The cancer genome atlas: creating lasting value beyond its data.
\newblock \emph{Cell}, 173\penalty0 (2):\penalty0 283--285, 2018.

\bibitem[Li et~al.(2024{\natexlab{a}})Li, Wong, Zhang, Usuyama, Liu, Yang, Naumann, Poon, and Gao]{li2024llava}
Chunyuan Li, Cliff Wong, Sheng Zhang, Naoto Usuyama, Haotian Liu, Jianwei Yang, Tristan Naumann, Hoifung Poon, and Jianfeng Gao.
\newblock Llava-med: Training a large language-and-vision assistant for biomedicine in one day.
\newblock \emph{Advances in Neural Information Processing Systems}, 36, 2024{\natexlab{a}}.

\bibitem[Li et~al.(2024{\natexlab{b}})Li, Chen, Chen, Yu, Yang, Wang, Ding, and Han]{li2024generalizable}
Hao Li, Ying Chen, Yifei Chen, Rongshan Yu, Wenxian Yang, Liansheng Wang, Bowen Ding, and Yuchen Han.
\newblock Generalizable whole slide image classification with fine-grained visual-semantic interaction.
\newblock In \emph{Proceedings of the IEEE/CVF Conference on Computer Vision and Pattern Recognition}, pages 11398--11407, 2024{\natexlab{b}}.

\bibitem[Lu et~al.(2021)Lu, Chen, Williamson, Zhao, Shady, Lipkova, and Mahmood]{lu2021ai}
Ming~Y Lu, Tiffany~Y Chen, Drew~FK Williamson, Melissa Zhao, Maha Shady, Jana Lipkova, and Faisal Mahmood.
\newblock Ai-based pathology predicts origins for cancers of unknown primary.
\newblock \emph{Nature}, 594\penalty0 (7861):\penalty0 106--110, 2021.

\bibitem[Lu et~al.(2024{\natexlab{a}})Lu, Chen, Williamson, Chen, Liang, Ding, Jaume, Odintsov, Le, Gerber, et~al.]{lu2024visual}
Ming~Y Lu, Bowen Chen, Drew~FK Williamson, Richard~J Chen, Ivy Liang, Tong Ding, Guillaume Jaume, Igor Odintsov, Long~Phi Le, Georg Gerber, et~al.
\newblock A visual-language foundation model for computational pathology.
\newblock \emph{Nature Medicine}, 30\penalty0 (3):\penalty0 863--874, 2024{\natexlab{a}}.

\bibitem[Lu et~al.(2024{\natexlab{b}})Lu, Chen, Williamson, Chen, Zhao, Chow, Ikemura, Kim, Pouli, Patel, et~al.]{lu2024multimodal}
Ming~Y Lu, Bowen Chen, Drew~FK Williamson, Richard~J Chen, Melissa Zhao, Aaron~K Chow, Kenji Ikemura, Ahrong Kim, Dimitra Pouli, Ankush Patel, et~al.
\newblock A multimodal generative ai copilot for human pathology.
\newblock \emph{Nature}, pages 1--3, 2024{\natexlab{b}}.

\bibitem[Radford et~al.(2021)Radford, Kim, Hallacy, Ramesh, Goh, Agarwal, Sastry, Askell, Mishkin, Clark, et~al.]{radford2021learning}
Alec Radford, Jong~Wook Kim, Chris Hallacy, Aditya Ramesh, Gabriel Goh, Sandhini Agarwal, Girish Sastry, Amanda Askell, Pamela Mishkin, Jack Clark, et~al.
\newblock Learning transferable visual models from natural language supervision.
\newblock In \emph{International conference on machine learning}, pages 8748--8763. PMLR, 2021.

\bibitem[Seyfioglu et~al.(2024)Seyfioglu, Ikezogwo, Ghezloo, Krishna, and Shapiro]{seyfioglu2024quilt}
Mehmet~Saygin Seyfioglu, Wisdom~O Ikezogwo, Fatemeh Ghezloo, Ranjay Krishna, and Linda Shapiro.
\newblock Quilt-llava: Visual instruction tuning by extracting localized narratives from open-source histopathology videos.
\newblock In \emph{Proceedings of the IEEE/CVF Conference on Computer Vision and Pattern Recognition}, pages 13183--13192, 2024.

\bibitem[Shao et~al.(2023)Shao, Chen, Bian, Zhang, Liu, and Zhang]{shao2023hvtsurv}
Zhuchen Shao, Yang Chen, Hao Bian, Jian Zhang, Guojun Liu, and Yongbing Zhang.
\newblock Hvtsurv: Hierarchical vision transformer for patient-level survival prediction from whole slide image.
\newblock In \emph{Proceedings of the AAAI Conference on Artificial Intelligence}, pages 2209--2217, 2023.

\bibitem[Song et~al.(2023)Song, Jaume, Williamson, Lu, Vaidya, Miller, and Mahmood]{song2023artificial}
Andrew~H Song, Guillaume Jaume, Drew~FK Williamson, Ming~Y Lu, Anurag Vaidya, Tiffany~R Miller, and Faisal Mahmood.
\newblock Artificial intelligence for digital and computational pathology.
\newblock \emph{Nature Reviews Bioengineering}, 1\penalty0 (12):\penalty0 930--949, 2023.

\bibitem[Spronck et~al.(2023)Spronck, Gelton, van Eekelen, Bogaerts, Tessier, van Rijthoven, van~der Woude, van~den Heuvel, Theelen, van~der Laak, et~al.]{spronck2023nnunet}
Joey Spronck, Thijs Gelton, Leander van Eekelen, Joep Bogaerts, Leslie Tessier, Mart van Rijthoven, Lieke van~der Woude, Michel van~den Heuvel, Willemijn Theelen, Jeroen van~der Laak, et~al.
\newblock nnunet meets pathology: bridging the gap for application to whole-slide images and computational biomarkers.
\newblock In \emph{Medical Imaging with Deep Learning}, 2023.

\bibitem[Sun et~al.(2024)Sun, Zhu, Zheng, Zhang, Sun, Shui, Zhang, Li, and Yang]{sun2024pathasst}
Yuxuan Sun, Chenglu Zhu, Sunyi Zheng, Kai Zhang, Lin Sun, Zhongyi Shui, Yunlong Zhang, Honglin Li, and Lin Yang.
\newblock Pathasst: A generative foundation ai assistant towards artificial general intelligence of pathology.
\newblock In \emph{Proceedings of the AAAI Conference on Artificial Intelligence}, pages 5034--5042, 2024.

\bibitem[Vorontsov et~al.(2024)Vorontsov, Bozkurt, Casson, Shaikovski, Zelechowski, Severson, Zimmermann, Hall, Tenenholtz, Fusi, et~al.]{vorontsov2024foundation}
Eugene Vorontsov, Alican Bozkurt, Adam Casson, George Shaikovski, Michal Zelechowski, Kristen Severson, Eric Zimmermann, James Hall, Neil Tenenholtz, Nicolo Fusi, et~al.
\newblock A foundation model for clinical-grade computational pathology and rare cancers detection.
\newblock \emph{Nature Medicine}, pages 1--12, 2024.

\bibitem[Wang et~al.(2024)Wang, Zhao, Marostica, Yuan, Jin, Zhang, Li, Tang, Wang, Li, et~al.]{wang2024pathology}
Xiyue Wang, Junhan Zhao, Eliana Marostica, Wei Yuan, Jietian Jin, Jiayu Zhang, Ruijiang Li, Hongping Tang, Kanran Wang, Yu Li, et~al.
\newblock A pathology foundation model for cancer diagnosis and prognosis prediction.
\newblock \emph{Nature}, pages 1--9, 2024.

\bibitem[Xu et~al.(2021)Xu, Zhu, Tang, Wang, Zhang, Li, Jiang, Shi, Liu, and Jin]{xu2021predicting}
Feng Xu, Chuang Zhu, Wenqi Tang, Ying Wang, Yu Zhang, Jie Li, Hongchuan Jiang, Zhongyue Shi, Jun Liu, and Mulan Jin.
\newblock Predicting axillary lymph node metastasis in early breast cancer using deep learning on primary tumor biopsy slides.
\newblock \emph{Frontiers in oncology}, 11:\penalty0 759007, 2021.

\bibitem[Xu et~al.(2024{\natexlab{a}})Xu, Usuyama, Bagga, Zhang, Rao, Naumann, Wong, Gero, Gonz{\'a}lez, Gu, et~al.]{xu2024whole}
Hanwen Xu, Naoto Usuyama, Jaspreet Bagga, Sheng Zhang, Rajesh Rao, Tristan Naumann, Cliff Wong, Zelalem Gero, Javier Gonz{\'a}lez, Yu Gu, et~al.
\newblock A whole-slide foundation model for digital pathology from real-world data.
\newblock \emph{Nature}, pages 1--8, 2024{\natexlab{a}}.

\bibitem[Xu et~al.(2024{\natexlab{b}})Xu, Wang, Zhou, Ma, Yang, Lin, Wang, Wang, Liang, Han, et~al.]{xu2024multimodal}
Yingxue Xu, Yihui Wang, Fengtao Zhou, Jiabo Ma, Shu Yang, Huangjing Lin, Xin Wang, Jiguang Wang, Li Liang, Anjia Han, et~al.
\newblock A multimodal knowledge-enhanced whole-slide pathology foundation model.
\newblock \emph{arXiv preprint arXiv:2407.15362}, 2024{\natexlab{b}}.

\bibitem[Yang et~al.(2024)Yang, Yang, Hui, Zheng, Yu, Zhou, Li, Li, Liu, Huang, et~al.]{yang2024qwen2}
An Yang, Baosong Yang, Binyuan Hui, Bo Zheng, Bowen Yu, Chang Zhou, Chengpeng Li, Chengyuan Li, Dayiheng Liu, Fei Huang, et~al.
\newblock Qwen2 technical report.
\newblock \emph{arXiv preprint arXiv:2407.10671}, 2024.

\bibitem[Zhang et~al.(2023)Zhang, He, Wu, Huang, Qin, Wang, Ye, Huang, Liao, Chen, et~al.]{zhang2023pathnarratives}
Heyu Zhang, Yan He, Xiaomin Wu, Peixiang Huang, Wenkang Qin, Fan Wang, Juxiang Ye, Xirui Huang, Yanfang Liao, Hang Chen, et~al.
\newblock Pathnarratives: Data annotation for pathological human-ai collaborative diagnosis.
\newblock \emph{Frontiers in Medicine}, 9:\penalty0 1070072, 2023.

\bibitem[Zhou and Chen(2023)]{zhou2023cross}
Fengtao Zhou and Hao Chen.
\newblock Cross-modal translation and alignment for survival analysis.
\newblock In \emph{Proceedings of the IEEE/CVF International Conference on Computer Vision}, pages 21485--21494, 2023.

\end{thebibliography}
}

\clearpage
\setcounter{page}{1}
\maketitlesupplementary

\section{SlideInstruction and SlideBench}

\subsection{Data Source}
In this section, we present the sources of the constructed SlideInstruction and SlideBench, which are derived from ten TCGA datasets as well as the BCNB challenge dataset. The~\cref{tb: dataset source} provides a detailed overview of the specific number of WSIs. 

\begin{table}[h]
\caption{Datasets statistics}
\label{tb: dataset source}
\begin{center}
\setlength\tabcolsep{2pt}
\resizebox{\linewidth}{!}{
\begin{tabular}{lccccc}
\toprule
Dataset  & WSIs & Report & Organ & Purpose
\\ \hline
TCGA-BRCA  & 1068  & \ding{51}  & Breast  & Train, Test \\
TCGA-LGG  & 783  & \ding{51}  & Brain  & Train, Test \\
TCGA-GBM  & 513  & \ding{51}  & Brain  & Train, Test \\
TCGA-LUAD  & 506  & \ding{51}  & Lung  & Train, Test \\
TCGA-LUSC  & 474 & \ding{51}  & Lung  & Train, Test \\
TCGA-HNSC  & 464  & \ding{51}  & Head and Neck  & Train, Test \\
TCGA-BLCA  & 424  & \ding{51}  & Bladder  & Train, Test \\
TCGA-COAD  & 419  & \ding{51}  & Colon  & Train, Test \\
TCGA-READ  & 157  & \ding{51}  & Rectum  & Train, Test \\
TCGA-SKCM  & 107  & \ding{51}  & Skin  & Train, Test \\ \hline
BCNB  &  1058 &  \ding{55} & Breast  & Test\\
\bottomrule
\end{tabular}
}
\end{center}
\end{table}

\subsection{Data Statistics}
We have compiled statistics on the number of VQA instances for each category within SlideBench VQA (TCGA) in ~\cref{tb: SlideBench-VQA (TCGA) num}. Each subcategory contains over 500 VQA instances, ensuring a robust representation across all areas, which supports comprehensive model evaluation and facilitates in-depth performance analysis.
We provide an overview of the sample sizes and detailed original label information for the seven classification tasks within the BCNB dataset in ~\cref{tb: BCNB data}.

\begin{table}[t]
\centering
\caption{The number of VQA corresponding to each category in SlideBench-VQA (TCGA).}
\setlength\tabcolsep{2.5pt}
\resizebox{\linewidth}{!}{
\begin{tabular}{ccc}
\toprule
Broad Category              & Narrow Catgory                      & Number  \\ \toprule
\multirow{5}{*}{Microscopy} & Tissue Architecture and Arrangement & 696  \\
                            & Tumor Characteristics               & 562  \\
                            & Cytomorphological Characteristics   & 601  \\
                            & Histopathological Changes           & 633  \\\hline
\multirow{6}{*}{Diagnosis}  & Disease Detection                   & 581  \\
                            & Disease Classification              & 532  \\
                            & Staging                             & 671  \\
                            & Grading                             & 601  \\
                            & Differential Diagnosis              & 586  \\ \hline
\multirow{5}{*}{Clinical}   & Treatment Guidance                  & 597  \\
                            & Biomarker Analysis                  & 502  \\
                            & Risk Factors                        & 591  \\
                            & Prognostic Assessment               & 674  \\  \bottomrule
\end{tabular}
}
\label{tb: SlideBench-VQA (TCGA) num}
\end{table}

\begin{table}[t]
\centering
\caption{The number and options of VQA corresponding to each task in SlideBench-VQA (BCNB).}
\setlength\tabcolsep{2.5pt}
\resizebox{\linewidth}{!}{
\begin{tabular}{lcl}
\toprule
Task                 & Number  & Option                                                               \\ \toprule
ER Status                  & 1058 & Postive / Negative    \\ \hline
HR Status                  & 1058 & Postive / Negative    \\ \hline
HER2 Status                 & 1058 & Postive / Negative    \\ \hline
HER2 Expression      & 1058 & 0 / 1+ / 2+ / 3+   \\ \hline
Histological Grading & 926  & 1 / 2 / 3           \\ \hline
Molecular Subtype    & 1058 & \makecell[l]{Luminal A / Luminal B \\/ HER2(+) / Triple negative}                   \\ \hline
Tumor Type               & 1058 & \makecell[l]{Invasive ductal carcinoma\\ / Invasive lobular carcinoma\\ / Other Type} \\ 
\bottomrule
\end{tabular}
}
\label{tb: BCNB data}
\end{table}

\subsection{Curation Scope and Prompt}

In this section, we illustrate the various dimensions of VQAs in SlideInstruction and SlideBench, ensuring comprehensive coverage of diverse pathological scenarios. This includes 3 broad categories and 13 narrow categories. Below are the contents for each category, which help to delineate their scope and meaning, thereby enabling GPT to extract high-quality question-answer pairs more effectively.

\begin{figure*}[t]
  \centering
  \includegraphics[width=\linewidth]{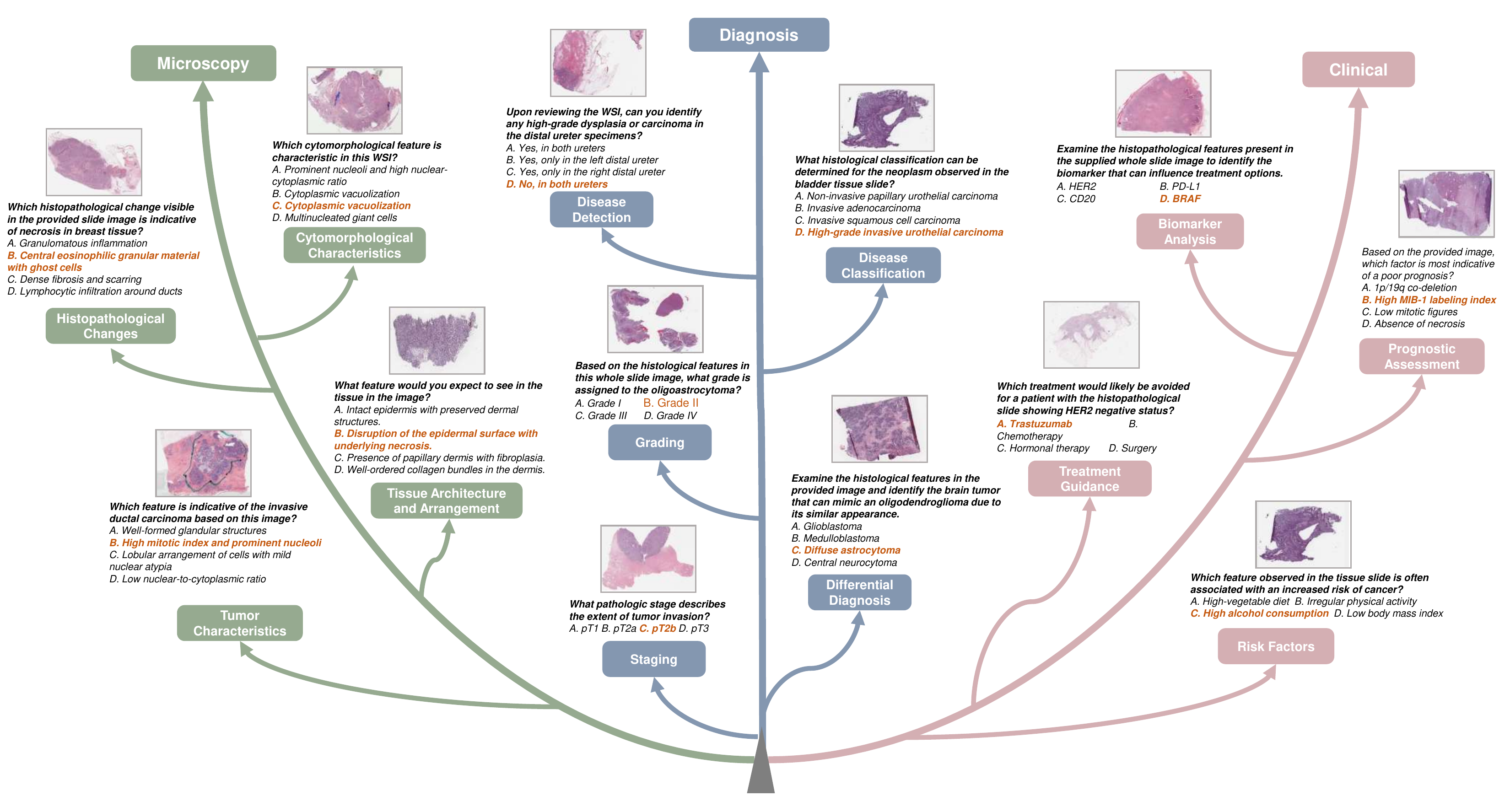}
  \caption{Examples of generated structural VQAs in pathology across Microscopy, Diagnosis, and Clinical scenarios.}
  \label{fig:vqa-structure}
\end{figure*}

\subsubsection{Scope}

\paragraph{Microscopy} This category involves assessing the ability to generate microscopy descriptions of pathology images, focusing on clinically relevant features:

\begin{itemize}[leftmargin=1em]
    \item{Tissue Architecture and Arrangement}: Questions in this category should evaluate the understanding of overall tissue structure and spatial organization within a histological section. 
    
    \item{Cytomorphological Characteristics}: These questions should focus on the detailed description of individual cell morphology, including nuclear and cytoplasmic features.

    \item {Tumor Characteristics}: Questions under this category should assess the ability to identify and describe features specific to tumors, such as tumor differentiation, invasion, and specific patterns associated with different types of tumors.
    
    \item {Histopathological Changes}: This category should include questions that evaluate the recognition and description of pathological changes in tissue, such as necrosis, inflammation, fibrosis, and other alterations that indicate disease processes.
\end{itemize}

\paragraph{Diagnosis} This category tests the ability of models to suggest a reasonable diagnosis based on histological images and relevant clinical context:

\begin{itemize}[leftmargin=1em]
    \item Disease Detection: Questions in this category should evaluate the model's ability to identify the presence or absence of a disease based on histological features and clinical information.
    
    \item Disease Classification: These questions should focus on distinguishing between different types or subtypes of diseases, assessing the model’s capability to classify conditions accurately based on morphological and histopathological criteria.
    
    \item Grading: Questions under this category should assess the model’s ability to determine the grade of a disease, particularly tumors, based on the degree of differentiation and cellular atypia observed in histological images.
    
    \item Staging: This category should include questions that evaluate the ability to assign a stage to a disease, particularly in oncology, by assessing the extent of disease spread and involvement of surrounding tissues or organs.
    
    \item Differential Diagnosis: Questions should test the model’s ability to provide a differential diagnosis, distinguishing between multiple potential conditions that may present with similar histological and clinical features.
\end{itemize}

\paragraph{Clinical} This category tests the ability of models to retrieve and apply clinically relevant background knowledge about diseases:

\begin{itemize}[leftmargin=1em]
    \item Treatment Guidance: Questions in this category should assess the model's ability to recommend appropriate treatment options based on the disease in question, considering factors such as disease stage, patient demographics, and any specific clinical guidelines.
    
    \item Prognostic Assessment: These questions should focus on evaluating the model's ability to predict the likely course and outcome of a disease, including survival rates, potential complications, and long-term outcomes based on clinical and pathological data.
    
    \item Risk Factors: Questions under this category should test the model's knowledge of risk factors associated with specific diseases, including genetic, environmental, and lifestyle factors that may influence disease development or progression.
    
    \item Biomarker Analysis: This category should include questions that evaluate the ability to identify and interpret biomarkers relevant to the diagnosis, prognosis, or treatment of diseases, emphasizing their role in personalized medicine and targeted therapy.
\end{itemize}

\subsubsection{Designed Prompts}
\label{ap:designed prompt}
\paragraph{Report Cleaning Prompt.}

\paragraph{Report Cleaning Prompt.}
The prompt used to clean up the report from the original TCGA report is represented in~\cref{tab:caption_prompt}. This process effectively eliminates extraneous noise from the report, thereby establishing a more solid foundation for caption and QA pairs generation.

\begin{table}[h]
\caption{Prompts for report clean and caption generation.}
\centering
\begin{minipage}{\columnwidth}
\vspace{0mm}    
\centering
\begin{tcolorbox} 
\raggedright
\small
\paragraph{[Report Clean Prompt]}
This is the content from the pathology report. Please remove some redundant irrelevant information from the original report, such as technical details of pathology department procedures, Symbols unrelated to the pathological report, specimen handling and processing information, redundant administrative or legal statements, and some repeated information. Show me the cleaned report content. \newline

\paragraph{[Caption Generation Prompt]}
Based on the above pathological report content, generate a detailed paragraph that summarizes the essential pathological findings. The paragraph should include key information such as the diagnosis, tumor characteristics, margin status, lymph node involvement, and other relevant pathological findings. The summary should not mention the source being a report and should exclude any specific sizes or measurements. The paragraph should be written in a clear and cohesive manner, covering all important points without unnecessary details.

\end{tcolorbox}
\label{tab:caption_prompt}
\end{minipage}
\end{table}

\paragraph{Caption Generation Prompt.} 
The prompt used for caption generation from the refined report is detailed in~\cref{tab:caption_prompt}, ensuring that the generated caption effectively captures essential summarized information in report.

\paragraph{Question-Answers Generation Prompt.} The prompt used to extract QA from reports mainly consist of 4 parts (\textit{i.e.}, \textless Cleaned Report\textgreater + System Prompt + Objective Prompt + General Prompt), and the detailed content of each part is illustrated in~\cref{tab:QA_prompt}

\begin{table}[h]
\caption{Question-Answers generation prompts, including system prompt, general prompt and objective prompt.}
\centering
\begin{minipage}{\columnwidth}
\vspace{0mm}    
\centering
\begin{tcolorbox} 
\raggedright
\small
\paragraph{[System Prompt]}
You are an AI assistant proficient in digital pathology. You will receive a pathology report for whole slide images. \newline
\paragraph{[General Prompt]}
Based on the above pathological report content, your task is to use the provided information, create 2 multi-choice questions amd 2 short-answer questions for each narrow category. The design question should be able to be answered based on the content of the image. Design medical questions very carefully and only ask questions when you are sure of the answer.  Answers should be specific and avoid ambiguity. When generating questions, it is necessary to indicate their broad category and narrow category.  For multi-choice questions, you should (1) “question type” is “multi-choice questions”.  (2) Provide the options and answer and reasoning. Provide four answer choices (A, B, C, and D), ensuring that one choice is correct and the other three are plausible but incorrect.\ (3) Aim to include one answer that is incorrect but very similar to the correct one to increase the difficulty level.  For short-answer questions: (1) “question type” is “short-answer questions”. (2) Generating questions with different content from multiple-choice questions. For all questions: (1) Do not mention that the information source is report in “question”, “anwser”. (2) Return JSON format in {“question type”: xxx, “question”: xxx, “options”: [], “answer”: xxx, “broad category”: xxx, “narrow category”: xxx} for each question. The “options” section is empty for short-answer questions. \newline
\paragraph{[Objective Prompt]}
Definition of Broad Category and its corresponding Narrow Categories. ``
The required broad category is Microscopy, which involves assessing the ability to generate microscopy descriptions of pathology images, focusing on clinically relevant features. For the narrow category:  Tissue Architecture and Arrangement: Questions should evaluate the understanding of overall tissue structure and spatial organization within a histological section.''

\end{tcolorbox}
\label{tab:QA_prompt}
\end{minipage}
\end{table}

\paragraph{Label Transformation Prompt.} The prompt for transforming BCNB dataset is illustrated in~\cref{tab:label_transform_prompt}. We employ GPT to transform individual labels into a question-answer format based on the task type and corresponding classification labels, facilitating the testing of MLLM. For instance, in the context of a tumor type classification task, \textless Task\textgreater represents ``Tumor Type'', while \textless label 1\textgreater, \textless label 2\textgreater, and \textless label 3\textgreater are ``Invasive ductal carcinoma'', ``Invasive lobular carcinoma'', and ``Other Type'', respectively, enabling the generation of relevant QA pairs.

\begin{table}[h]
\caption{Prompt for Converting Labels into QA Pairs}
\centering
\begin{minipage}{\columnwidth}
\vspace{0mm}    
\centering
\begin{tcolorbox} 
\raggedright
\small
\paragraph{[Label Transformation Prompt]}
Please create prompts for pathology image classification tasks concerning \textless Task\textgreater, transforming traditional labels into a multi-choice question-and-answer format. The original labels include \textless label 1\textgreater, \textless label 2\textgreater, ...
\end{tcolorbox}
\label{tab:label_transform_prompt}
\end{minipage}
\end{table}

\subsection{Multimodal Dataset Comparsion}
Recently, several multimodal pathology datasets have been introduced for pathology applications. However, these datasets are often constrained in both scope and scale, as they primarily focus on either patch-level analysis or limited available data. In contrast, our proposed SlideInstruction and SlideBench, provided as open-source resources, significantly expand the dataset size while enhancing its versatility, as shown in~\cref{tb:dataset-compare}.

\begin{table}[h]
    \centering
    \caption{Comparisons of our datasets with other pathology datasets.}
    \label{tab:related_bench_comparsion}
    \resizebox{\linewidth}{!}{
    \begin{tabular}{lccccccc}
    \toprule
    Dataset & Level & Data Type & Curation Type & Scope & Number  & Availability \\
    \midrule
    PathChat~\citep{lu2024multimodal} & Patch & Patch and Q/A pairs & Human+GPT & - & 257,004 & \ding{55} \\
    Quilt-Instruct~\citep{seyfioglu2024quilt} & Patch & Patch and Q/A pairs & GPT & - & 107,131 & \ding{51} \\ \hline
    WSI-VQA~\citep{chen2024wsi}   & Slide  & WSI and Q/A pairs &  GPT      & - & 8,672   & \ding{51} \\
    PathText~\citep{chen2023mi} &  Slide & WSI-Caption pairs&  GPT     & - & 9,009     & \ding{51} \\
    HistGen~\citep{guo2024histgen} &  Slide & WSI-Reports pairs &  GPT     & - & 7,753      & \ding{51} \\
    Prov-Path~\citep{xu2024whole}  & Slide & WSI-Reports pairs & GPT & - & 17,383  & \ding{55} \\ 
    CR-PathNarratives~\citep{zhang2023pathnarratives} &  Slide & \makecell[c]{WSIs with \\annotations} & Human & - & 174 & \ding{55} \\ 
    PathAlign~\citep{ahmed2024pathalign} &  Slide & WSI-Reports pairs & Human & - & 354,089 & \ding{55} \\ 
    \hline
    Our SlideInstruction    &  Slide &  WSI and Q/A pairs & GPT   &  13 & 179,935   & \ding{51} \\
    Our SlideBench &  Slide   & WSI and Q/A pairs  & Human+GPT   &  13 & 15,835   & \ding{51} \\
    \bottomrule
    \end{tabular}}
    \label{tb:dataset-compare}
\end{table}

\section{Experiment}

\subsection{Computational Cost Analysis}
To evaluate the computational cost of our model architecture, we measured both the inference time and GPU memory consumption throughout the entire pipeline. This pipeline includes the patch-level encoder, slide-level encoder, multimodal projector module, and large language model, all executed on an A100 GPU. After extracting the local and global features of WSIs, the average response time was within 1 second, and GPU memory consumption was approximately 27 GB.
The inference time and GPU memory consumption remained well within acceptable limits for gigapixel whole slide images.

\subsection{Implementation Details}
We preprocessed each WSI by segmenting it into 224 × 224 nonoverlapping patches at a 20× magnification level, excluding background regions. We implemented our model using the Xtuner~\citep{contributors2023xtuner} toolkit and trained it across two stages on 8 × NVIDIA A100 GPUs. The training process consists of an alignment phase followed by instruction fine-tuning: 
Stage 1: We freeze the LLM and train the Projection and Slide Encoder with WSI-caption data for 3 epochs, using a learning rate of 0.001. 
Stage 2: We unfreeze the LLM, Slide Encoder, and Projection, training the model on WSI instruction-following data for 1 epoch, with a learning rate of 0.00002. 
Both stages are optimized using AdamW. 

\subsection{Ability Showcase}

\subsubsection{Captioning Ability}
\label{ap:Captioning case}
\begin{figure*}[h]
  \centering
  \includegraphics[width=\linewidth]{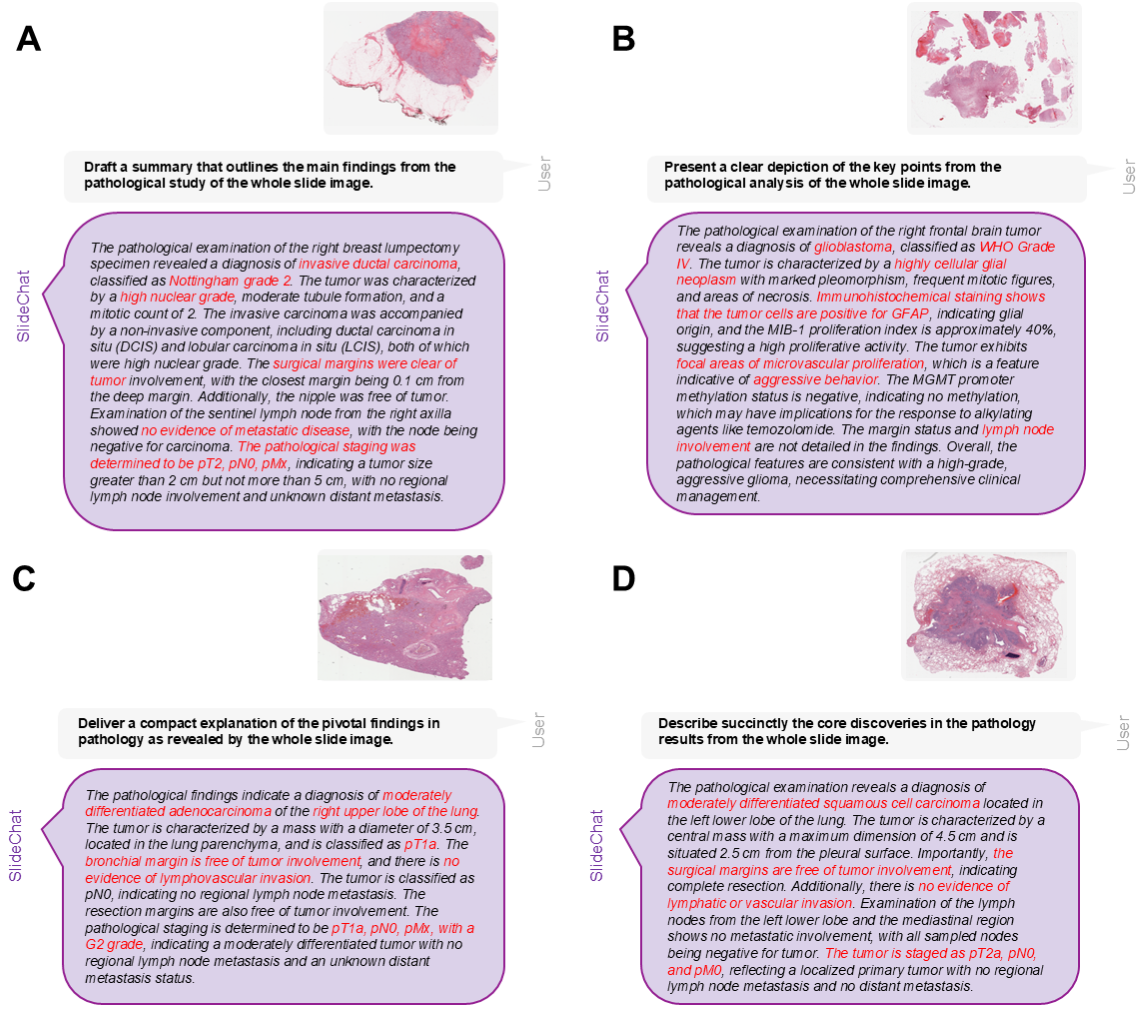}
  \caption{Demonstration of SlideChat's Capability in Whole-Slide Image Captioning.}
  \label{fig:caption-chat}
\end{figure*}

The examples shown in ~\cref{fig:caption-chat} illustrate the capability of our model, SlideChat, to effectively perform whole-slide image captioning tasks. SlideChat demonstrates its proficiency in generating detailed and contextually accurate summaries for complex pathological whole-slide images, accurately capturing key clinical findings and pathological features. Whether summarizing broad findings, explaining pivotal details, or highlighting core results, SlideChat showcases an advanced understanding of whole-slide images, providing concise yet informative reports that align with clinical terminology and expectations.

\subsubsection{VQA Ability}
\begin{figure*}[h]
  \centering
  \includegraphics[width=0.8\linewidth]{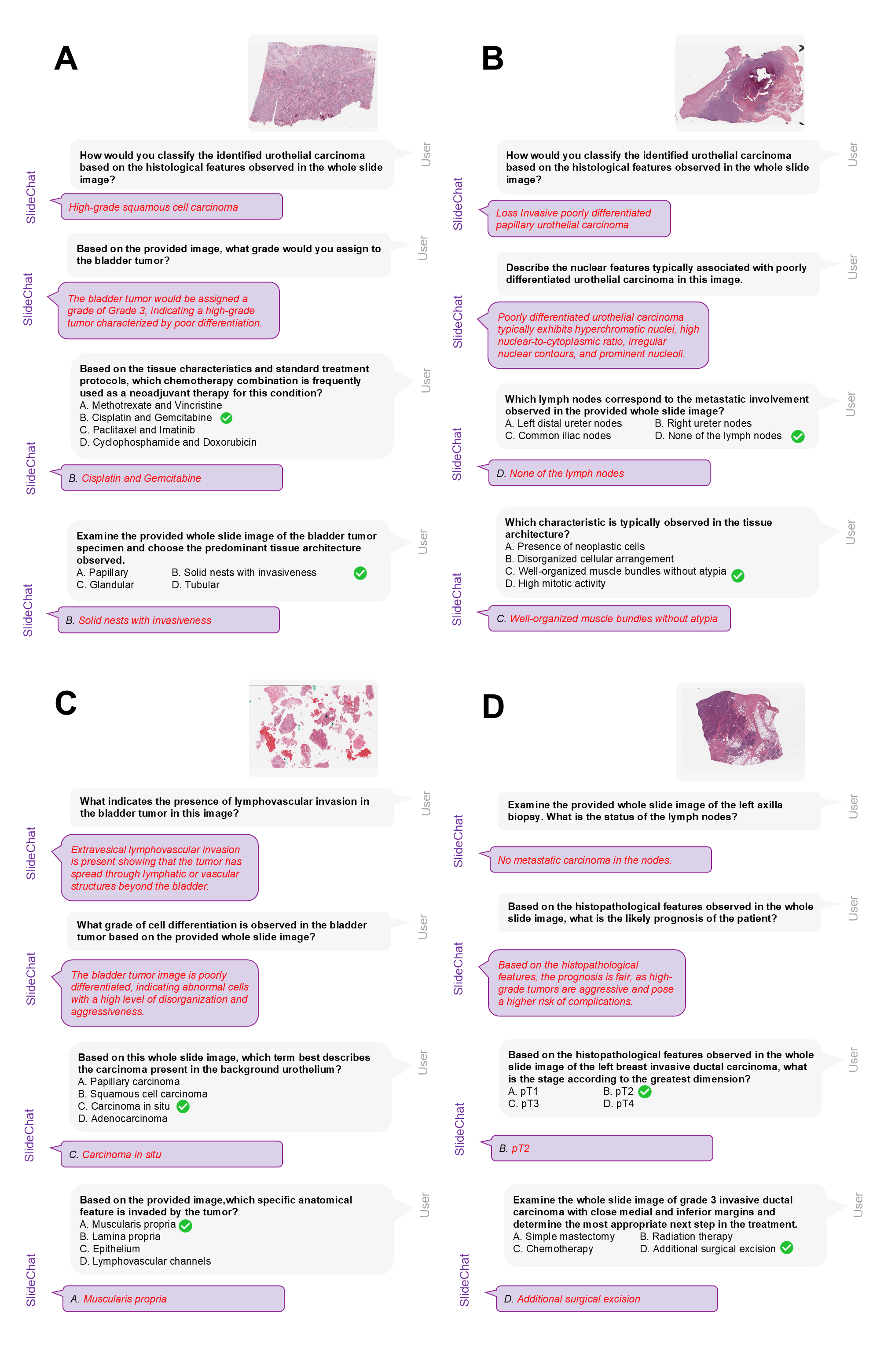}
  \caption{Demonstration of our SlideChat for answering various questions based on the WSI.}
  \label{fig:chat-case}
\end{figure*}

\cref{fig:chat-case} showcases the conversational examples of SlideChat, demonstrating its ability to accurately answer a range of questions based on WSIs, covering diverse aspects such as histological classifications, tumor grading, lymph node involvement, and treatment decisions. SlideChat effectively interprets complex pathological data, engages in nuanced question-and-answer exchanges, and delivers clinically relevant responses. This reflects its potential as an intelligent assistant capable of supporting pathologists in diagnostic decision-making by providing insightful, context-aware dialogue grounded in visual pathology data.

\subsubsection{Comparing Model Outputs}
\begin{figure*}[h]
  \centering
  \includegraphics[width=0.85\linewidth]{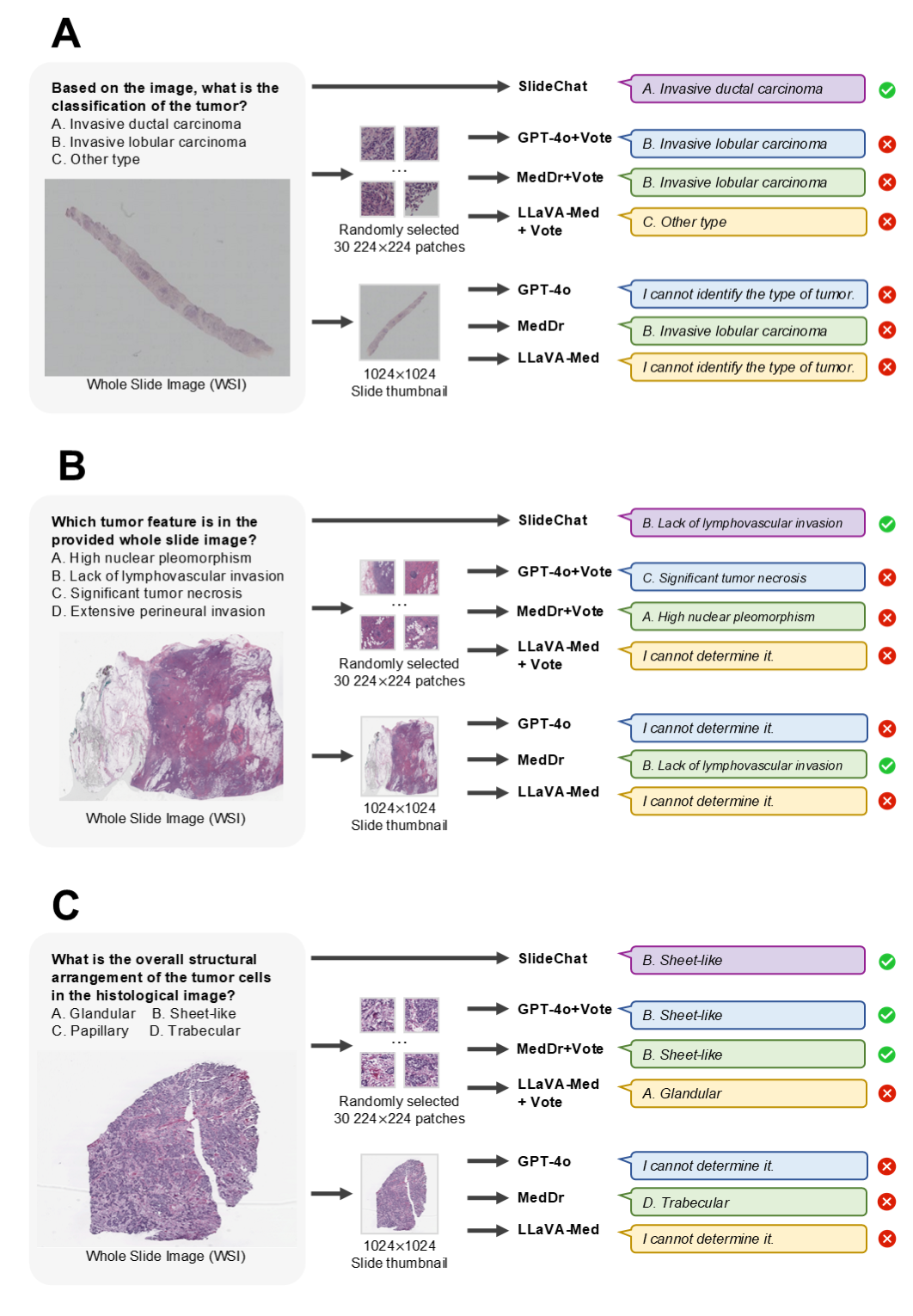}
  \caption{Comparing model outputs on SlideBench.}
  \label{fig:chat-comparsion}
\end{figure*}

\cref{fig:chat-comparsion} presents a comparative analysis of the outputs from SlideChat and other models within SlideBench. The examples illustrate SlideChat's remarkable capacity to precisely classify tumors, identify distinct histological features, and describe the structural organization of tumor cells from WSIs. SlideChat demonstrates a unique proficiency in capturing both local and global features—seamlessly integrating detailed microscopic characteristics with broader contextual understanding to deliver accurate and clinically meaningful interpretations. In contrast, existing models are limited to processing small pathology images, often yielding ambiguous or incorrect classifications. This underscores SlideChat's advanced capability in comprehending whole-slide images by incorporating both intricate details and a comprehensive visual perspective.

\subsection{Detailed Test Performance}

\begin{table*}[h]
\centering
\resizebox{0.9\linewidth}{!}{
\begin{tabular}{ccccccc}
\hline
&  & \multicolumn{4}{c}{SlideBench-VQA(TCGA)  \textbf{Microscopy}}   & \\ 
\cline{3-6}
\multirow{-2}{*}{Method} & \multirow{-2}{*}{Input}   & \begin{tabular}[c]{@{}c@{}}Tissue Architecture \\ and Arrangement\end{tabular} & \begin{tabular}[c]{@{}c@{}}Tumor \\ Characteristics\end{tabular} & \begin{tabular}[c]{@{}c@{}}Cytomorphological \\ Characteristics\end{tabular} & \begin{tabular}[c]{@{}c@{}}Histopathological \\ Changes\end{tabular} & \multirow{-2}{*}{Overall}                                \\ \hline
Random & \multirow{2}{*}{Text}  &  23.70  &  22.42  & 23.63  & 27.80  & 24.44  \\
GPT-4  &   &  40.83  & 40.28  & 41.71   & 37.46  & 39.62  \\ \hline
GPT-4o  & \multirow{4}{*}{Patch} & 65.94  & 66.20  & 60.10  & 59.23  & 62.89  \\
MedDr   &   & 75.04  & 75.78  & 70.10  & 72.23   & 73.30   \\
LLaVA-Med   &    & 50.04   & 40.63  & 40.38  & 56.95   & 47.34  \\ 
Quilt-LLaVA & & 65.26 &	54.04 &	50.66 &	59.55 &	57.76 \\ \hline
GPT-4o   & \multirow{4}{*}{Slide (T)}   & 37.07  & 38.76    & 39.93   & 37.60   & 38.28   \\
MedDr  &   & 71.58   & 71.27   & 69.87     & 69.05     & 70.48     \\
LLaVA-Med                &  & 51.80     & 45.02   & 36.27    & 49.01   & 45.82  \\ 
Quilt-LLaVA & & 53.59	& 45.37 &	43.09 &	53.24 &	49.12 \\ \hline
\rowcolor[HTML]{EFEFEF} 
SlideChat  & Slide   & \begin{tabular}[c]{@{}c@{}}88.07\\ \textcolor{blue}{(+13.03)}\end{tabular}                       & \begin{tabular}[c]{@{}c@{}}87.01\\ \textcolor{blue}{(+11.23)}\end{tabular}         & \begin{tabular}[c]{@{}c@{}}88.02\\ \textcolor{blue}{(+17.92)}\end{tabular}                     & \begin{tabular}[c]{@{}c@{}}87.36\\ \textcolor{blue}{(+15.13)}\end{tabular}             & \begin{tabular}[c]{@{}c@{}}87.64\\ \textcolor{blue}{(+14.34)}\end{tabular} \\ \hline
\end{tabular}
}
\end{table*}

\begin{table*}[h]
\centering
\resizebox{0.9\linewidth}{!}{
\begin{tabular}{cccccccc}
\hline
  &    & \multicolumn{5}{c}{SlideBench-VQA(TCGA)   \textbf{Diagnosis}}   \\ \cline{3-7}
\multirow{-2}{*}{Method} & \multirow{-2}{*}{Input}   & \begin{tabular}[c]{@{}c@{}}Disease \\ Detection\end{tabular} & \begin{tabular}[c]{@{}c@{}}Disease \\ Classification\end{tabular} & Staging             & Grading   & \begin{tabular}[c]{@{}c@{}}Differential \\ Diagnosis\end{tabular} & \multirow{-2}{*}{Overall}    \\ \hline
Random   & \multirow{2}{*}{Text}  & 25.82  & 24.06   & 24.14   & 26.12    & 24.40   & 24.91   \\
GPT-4    &   & 27.12  & 31.07  & 22.27    & 27.45   & 38.70  & 29.09   \\ \hline
GPT-4o   &  \multirow{4}{*}{Patch} & 50.27   & 55.94    & 39.94   & 39.66   & 49.66   & 46.69     \\
MedDr    &   & 59.11  & 61.11    & 48.66  & 52.97    & 68.83   & 57.78    \\
LLaVA-Med    &   & 37.25    & 28.57    & 30.41   & 20.71    & 47.27    & 32.78    \\ 
Quilt-LLaVA & & 40.74	& 39.3 &	32.32 &	28.96 &	39.52 &	35.96 \\ \hline
GPT-4o   & \multirow{4}{*}{Slide (T)} & 22.95 & 26.76 & 18.06   & 21.06 & 27.82 & 23.10    \\
MedDr    &  & 54.29  & 56.40    & 48.66     & 43.52  & 61.61   & 52.47     \\
LLaVA-Med    &  & 27.87  & 25.19    & 24.07    & 24.96     & 36.18    & 27.58  \\
Quilt-LLaVA  & & 32.47 &	28.25 &	20.18 &	22.96 &	32.25 &	26.97  \\ \hline
\rowcolor[HTML]{EFEFEF} 
SlideChat                & Slide                     & \begin{tabular}[c]{@{}c@{}}80.90\\ \textcolor{blue}{(+21.79)}\end{tabular}      & \begin{tabular}[c]{@{}c@{}}76.12\\ \textcolor{blue}{(+15.01)}\end{tabular}          & \begin{tabular}[c]{@{}c@{}}68.41\\ \textcolor{blue}{(+19.75)}\end{tabular} & \begin{tabular}[c]{@{}c@{}}68.39\\ \textcolor{blue}{(+15.42)}\end{tabular} & \begin{tabular}[c]{@{}c@{}}73.72\\ \textcolor{blue}{(+4.89)}\end{tabular}           & \begin{tabular}[c]{@{}c@{}}73.27\\ \textcolor{blue}{(+15.49)}\end{tabular} \\ \hline
\end{tabular}
}
\end{table*}

\begin{table*}[h]
\centering
\resizebox{0.8\linewidth}{!}{
\begin{tabular}{ccccccc}
\hline
  &   & \multicolumn{4}{c}{SlideBench-VQA(TCGA)    \textbf{Clinical}}   &  \\ \cline{3-6}
\multirow{-2}{*}{Method} & \multirow{-2}{*}{Input}   & \begin{tabular}[c]{@{}c@{}}Treatment \\ Guidance\end{tabular} & \begin{tabular}[c]{@{}c@{}}Biomarker \\ Analysis\end{tabular} & \begin{tabular}[c]{@{}c@{}}Risk \\ Factors\end{tabular} & \begin{tabular}[c]{@{}c@{}}Prognostic \\ Assessment\end{tabular} & \multirow{-2}{*}{Overall}     \\ \hline
Random    &  \multirow{2}{*}{Text}    & 23.62  & 31.87   & 24.36  & 24.33  & 26.44  \\
GPT-4     &    & 49.98   & 44.63  & 46.46  & 39.64   & 45.00 \\ \hline
GPT-4o   & \multirow{4}{*}{Patch}   & 64.18   & 57.99    & 76.99  & 66.64    & 66.77 \\
MedDr    &   & 74.18   & 82.99    & 82.43  & 60.66    & 74.25\\
LLaVA-Med  &   & 62.04   & 53.98   & 53.04  & 26.54   & 47.96   \\ 
Quilt-LLaVA & & 64.79 &	40.42 &	63.40 &	43.06 &	53.07	\\ \hline
GPT-4o   &  \multirow{4}{*}{Slide (T)} & 50.00   & 50.08    & 44.16    & 32.64     & 43.42    \\
MedDr   &   & 71.43    & 84.51     & 78.92   & 60.24     & 72.80     \\
LLaVA-Med                &  & 50.50   & 48.01    & 48.90    & 19.88    & 40.84    \\ 
Quilt-LLaVA & & 47.38 &	32.93 &	47.71 &	48.64 &	44.76	\\ \hline
\rowcolor[HTML]{EFEFEF} 
SlideChat                & Slide                     & \begin{tabular}[c]{@{}c@{}}83.42\\ \textcolor{blue}{(+9.24)}\end{tabular}       & \begin{tabular}[c]{@{}c@{}}89.04\\ \textcolor{blue}{(+4.53)}\end{tabular}       & \begin{tabular}[c]{@{}c@{}}91.71\\ \textcolor{blue}{(+9.28)}\end{tabular} & \begin{tabular}[c]{@{}c@{}}74.93\\ \textcolor{blue}{(+8.29)}\end{tabular}          & \begin{tabular}[c]{@{}c@{}}84.26\\ \textcolor{blue}{(+10.01)}\end{tabular} \\ \hline
\end{tabular}
}
\end{table*}

\begin{table*}[h]
\centering
\resizebox{0.9\linewidth}{!}{
\begin{tabular}{cccccccccc}
\hline
\multicolumn{1}{l}{} & \multicolumn{1}{l}{}    & \multicolumn{7}{c}{SlideBench-VQA(BCNB)}    & \multicolumn{1}{l}{} \\ \cline{3-9}
\multicolumn{1}{l}{\multirow{-2}{*}{Method}} & \multicolumn{1}{l}{\multirow{-2}{*}{Input}} & \begin{tabular}[c]{@{}c@{}}Tumor \\ Type\end{tabular} & \begin{tabular}[c]{@{}c@{}}ER \\ Type\end{tabular}   & \begin{tabular}[c]{@{}c@{}}PR \\ Type\end{tabular}  & \begin{tabular}[c]{@{}c@{}}HER2 \\ Type\end{tabular}     & \begin{tabular}[c]{@{}c@{}}HER2 \\ Expression\end{tabular} & \begin{tabular}[c]{@{}c@{}}Histological \\ Grading\end{tabular} & \begin{tabular}[c]{@{}c@{}}Molecular \\ Subtype\end{tabular} & \multicolumn{1}{l}{\multirow{-2}{*}{Overall}}            \\ \hline
Random   & \multirow{2}{*}{Text} & 23.82    & 24.48   & 25.05  & 25.05   & 24.39      & 24.41           & 23.63        & 24.40  \\
GPT-4    &  & 0        & 0       & 0      & 0       & 0          & 0               & 0            & 0     \\ \hline
GPT-4o   & \multirow{4}{*}{Patch}  & 34.69    & 77.50    & 63.51  & 36.95   & 23.95      & 28.63           & 23.15        & 41.43     \\
MedDr    &  & 45.46    & 23.53   & 25.99  & 71.81   & 22.73      & 30.28           & 15.49        & 33.67       \\
LLaVA-Med     &                     & 23.95    & 36.62   & 40.19  & 50.76   & 23.72      & 18.99           & 15.05        & 30.10  \\ 
Quilt-LLaVA & & 77.14	& 68.58 &	42.63 &	58.17 &	23.18 &	18.23 &	19.82 &	44.43 \\ \hline
GPT-4o   & \multirow{4}{*}{Slide (T)} & 0        & 0       & 0      & 0       & 0          & 0               & 0            & 0     \\
MedDr    &  & 28.92    & 45.84   & 25.71  & 72.68   & 20.65      & 29.96           & 23.88        & 35.48                                                    \\
LLaVA-Med  &                    & 0.01     & 0       & 0.01   & 0.02    & 0          & 0               & 0            & 0.01  \\ 
Quilt-LLaVA & & 67.41 &	66.73 &	36.58 &	62.67 &	15.97 &	22.89 &	16.27 &	41.55 \\ \hline
\rowcolor[HTML]{EFEFEF} 
SlideChat                                    & Slide   & \begin{tabular}[c]{@{}c@{}}90.17\\ \textcolor{blue}{(+44.71)}\end{tabular} & \begin{tabular}[c]{@{}c@{}}78.54\\ \textcolor{blue}{(+1.04)}\end{tabular} & \begin{tabular}[c]{@{}c@{}}68.81\\ \textcolor{blue}{(+5.3)}\end{tabular} & \begin{tabular}[c]{@{}c@{}}71.93\\ \textcolor{blue}{(-0.75)}\end{tabular} & \begin{tabular}[c]{@{}c@{}}25.05\\ \textcolor{blue}{(+0.66)}\end{tabular}    & \begin{tabular}[c]{@{}c@{}}23.11\\ \textcolor{blue}{(-7.17)}\end{tabular}         & \begin{tabular}[c]{@{}c@{}}17.49\\ \textcolor{blue}{(-6.39)}\end{tabular}      & \begin{tabular}[c]{@{}c@{}}54.14\\ \textcolor{blue}{(+12.71)}\end{tabular} \\ \hline
\end{tabular}
}
\end{table*}

\subsubsection{Performance on SlideBench-VQA (TCGA)}
The results presented in the tables demonstrate a comprehensive evaluation of SlideChat's performance on SlideBench-VQA (TCGA) in comparison to other existing models across microscopy, diagnosis, and clinical tasks. In microscopy, SlideChat significantly outperforms its counterparts, achieving a notable overall accuracy improvement of 14.34 points over the nearest model. This strong performance is consistent across sub-tasks, such as tissue architecture analysis, tumor characteristics identification, and cytomorphological assessment, showcasing SlideChat's advanced capability to analyze both detailed cellular structures and broader histopathological changes. In the diagnostic tasks, SlideChat also demonstrates superior accuracy, with an overall gain of 15.49 points, excelling in disease detection, classification, staging, grading, and differential diagnosis. The clinical analysis results further validate the model's strength, with SlideChat outperforming other methods by 10.01 points overall, particularly excelling in treatment guidance, biomarker analysis, and risk factor assessment. These results illustrate SlideChat's capability to seamlessly handle complex medical data and deliver reliable insights across multiple clinical and diagnostic domains, indicating its potential as a robust tool for comprehensive pathology analysis.

\subsubsection{Performance on SlideBench-VQA (BCNB)}
The evaluation of SlideChat on SlideBench-VQA (BCNB), a real-world dataset designed for zero-shot testing, further underscores its ability to generalize effectively to unseen data. SlideChat demonstrates an overall accuracy improvement of 12.71 points compared to other models, showcasing its ability to generalize well across diverse and complex breast cancer-related tasks. SlideChat's performance is particularly strong in identifying tumor type, ER status, PR status, and HER2 status, demonstrating a nuanced understanding of critical histopathological features. Nevertheless, in the more complex tasks of HER2 Expression, Histological Grading, and Molecular Subtype classification, SlideChat still exhibits potential for improvement, highlighting specific areas that warrant further refinement to enhance its overall performance.

\FloatBarrier

\FloatBarrier

\FloatBarrier


\end{document}